\documentclass{article}
\usepackage{PRIMEarxiv}
\usepackage{xcolor}
\usepackage{pifont}
\usepackage{tabularx}
\usepackage{array}
\usepackage{amsmath, amsfonts, amsthm, amssymb,epsfig}
\usepackage{authblk}
\usepackage[utf8]{inputenc} % allow utf-8 input
\usepackage[T1]{fontenc}    % use 8-bit T1 fonts
\usepackage{hyperref}       % hyperlinks
\usepackage{url}            % simple URL typesetting
\usepackage{booktabs}       % professional-quality tables
\usepackage{amsfonts}       % blackboard math symbols
\usepackage{nicefrac}       % compact symbols for 1/2, etc.
\usepackage{microtype}      % microtypography
\usepackage{lipsum}
\usepackage{fancyhdr}       % header
\usepackage{graphicx}       % graphics
\graphicspath{{media/}} 
\usepackage{subcaption}
\usepackage{algorithm}
\usepackage{algpseudocode}
\usepackage[absolute,overlay]{textpos}
\usepackage{caption}
\usepackage{bm}
\title{Video-based Surgical Skill Assessment Using Dynamics-and-Uncertainty-Aware Tree-based Gaussian Process Classifier
}

\author[1,*]{Arefeh Rezaei}
\author[1,*]{Mohammad Javad Ahmadi}
\author[2]{Amir Molaei}
\author[1]{Hamid D. Taghirad}

\affil[1]{Applied Robotics and AI Solutions (ARAS), Faculties of Electrical and Computer Engineering, K.N. Toosi University of Technology, Tehran, Iran}
\affil[2]{Gina Cody School of Engineering and Computer Science, Concordia University, Montreal, Quebec, Canada}

\DeclareUnicodeCharacter{202A}{}
\DeclareUnicodeCharacter{202C}{}
\begin{document}
% --- COVER PAGE (page 1) ---
\thispagestyle{empty}
\begingroup
\setlength{\parskip}{6pt}
\setlength{\parindent}{0pt}

\definecolor{coverAccent}{HTML}{0B6E99} % accent
\definecolor{coverSubtle}{HTML}{6B7280} % gray-600

\begin{center}
  % {\Large\textsf{Scientific Data}}\par
  % {\footnotesize\color{coverSubtle}\textsf{Data Descriptor}}\par
  % \vspace{5mm}
  {\color{coverAccent}\rule{\linewidth}{0.8pt}}\vspace{7mm}

  {\huge\bfseries Video-based Surgical Skill Assessment Using Dynamics-and-Uncertainty-Aware Tree-based Gaussian Process Classifier\par}
  \vspace{7mm}
% \textsuperscript{\ding{72}}   % ★ black star
% \textsuperscript{\dag}        % † dagger

  % author line: * = corresponding, † = co-corresponding
  {\normalsize
    Arefeh Rezaei$^{1}$\textsuperscript{\,\textsuperscript{\ding{72}}},
    Mohammad Javad Ahmadi$^{1}$\textsuperscript{\,\textsuperscript{\ding{72}}},
    Amir Molaei$^{2}$,
    {Hamid D. Taghirad}$^{1}$\textsuperscript{\,\textsuperscript{\ding{72}\ding{72}}}\par}

  \vspace{4mm}
    {\color{coverAccent}\rule{\linewidth}{0.8pt}}\vspace{7mm}

  % {\color{coverAccent}\rule{0.25\linewidth}{0.8pt}}\vspace{4mm}
\end{center}

{\small
\textbf{Affiliations}\par
\vspace{2mm}
$^{1}$ Applied Robotics and AI Solutions (ARAS), Faculties of Electrical and Computer Engineering, K.N. Toosi University of Technology, Tehran, Iran.\par
$^{2}$ Gina Cody School of Engineering and Computer Science, Concordia University, Montreal, Quebec, Canada.\par
}
% {\color{coverAccent}\rule{0.25\linewidth}{0.8pt}}\vspace{4mm}

\vspace{3mm}
{\small
\textbf{Author e-mails}\par
\vspace{1mm}
\begin{tabularx}{\textwidth}{@{}>{\raggedright\arraybackslash}p{4.6cm} >{\raggedright\arraybackslash}X@{}}
Arefeh Rezaei\textsuperscript{\,\textsuperscript{\ding{72}}} & \href{mailto:a_rezai@email.kntu.ac.ir}{a\_rezai@email.kntu.ac.ir} \\
Mohammad Javad Ahmadi\textsuperscript{\,\textsuperscript{\ding{72}}} & \href{mailto:mjahmadi@email.kntu.ac.ir}{mjahmadi@email.kntu.ac.ir} \\
Amir Molaei & \href{mailto:a_molaei@encs.concordia.ca}{a\_molaei@encs.concordia.ca} \\
Hamid D. Taghirad\textsuperscript{\,\textsuperscript{\ding{72}\ding{72}}} & \href{mailto:taghirad@kntu.ac.ir}{taghirad@kntu.ac.ir} \\
\end{tabularx}
}

% {\color{coverAccent}\rule{0.25\linewidth}{0.8pt}}\vspace{4mm}

\vspace{2mm}
{\small
\textsuperscript{\textsuperscript{\ding{72}}}\,These authors contributed equally to this work.\par
\textbf{Correspondence}: \textsuperscript{\textsuperscript{\ding{72}\ding{72}}}\,Hamid D. Taghirad (\href{mailto:taghirad@kntu.ac.ir}{taghirad@kntu.ac.ir}) \par
}

\vspace{1mm}
{\footnotesize
\textbf{Keywords:}\;
computer-assisted surgery; surgical skill assessment; representation flow; optical flow; Gaussian process; video classification.\par
}

% \vfill
% {\color{coverSubtle}\footnotesize Submitted to \textit{BMC Medical Informatics and Decision Making}.\par}
\endgroup
\clearpage
% --- END COVER PAGE ---

\maketitle

\begin{abstract}
The proposed pipeline integrates a representation-flow convolutional neural network with a dynamics- and uncertainty-aware tree-based Gaussian Process classifier. In this framework, latent motion dynamics are exploited both as discriminative representations and as a source of input uncertainty, enhancing robustness against temporal variations and abnormal motion transitions. The framework targets limited-data settings and combines motion-aware video features with a compact probabilistic classifier; its post-feature-extraction inference cost and classifier complexity are reported separately from the feature-extraction cost. To further improve classification performance, we introduce novel semantic-aware compound kernels that effectively capture semantic, flow, and dynamic information embedded in surgical video features. In addition, uncertainty-aware kernels are developed to strengthen the robustness and practical applicability of the compound kernel framework. The proposed method is evaluated on two benchmark datasets, namely the JIGSAWS and the Cataract-LMM (Capsulorhexis) datasets, through two documented dataset-specific instantiations of the same pipeline. Under the subject-independent LOUO protocol on JIGSAWS, the framework attains a mean accuracy of 96.90\%, while on Capsulorhexis it achieves a mean accuracy of 89.2\% under the repeated evaluation protocol, using 30 training videos per run. The published deep baselines for Capsulorhexis used approximately 140 training videos. Results under the within-subject LOSO protocol, on which this benchmark is known to saturate, are reported for comparability with prior work. Overall, the proposed pipeline provides an efficient and accurate framework for video-based surgical skill assessment.
\end{abstract}

% keywords can be removed
\keywords{computer-assisted surgery; surgical skill assessment; representation flow; optical flow; Gaussian process; video classification.}

\section{Introduction}
Surgical skill assessment is the process of evaluating a surgeon's technical ability and competence in performing surgical procedures to ensure patient safety and the quality of care delivered. Skill assessments play a crucial role in certifying aspiring surgeons~\cite{power2025automated}. Such assessments aid in efficient skill acquisition by providing specific guidance on areas that need improvement~\cite{FormativeandSummativeAssessmentNIU, hashemi2025video, power2025automated,kim2025generalizability,lam2022machine}.  The notion of automating the process of surgical skill evaluation has emerged as a promising solution~\cite{liu2025vision,lam2022machine}. This approach could save time and resources, while also enabling novice surgeons to train with increased autonomy (using a surgical simulator that provides automatic assessment and feedback), moving towards independence in surgical skill assessment. However, it is important to note that this progress is still in line with the current trend in surgical automation, which emphasizes augmented assistance rather than complete replacement of expert supervision. As a result, there has been a growing interest in the research community on automatic skill evaluation over the past decade.

In automated surgical skill assessment (ASSA), sensory data such as robot kinematics, tool motion, force, or video data are collected during surgical training. This data is subsequently analyzed to determine the trainee's surgical skill level, which can be expressed either numerically as the average score or categorically as novice, intermediate, or expert~\cite{liu2025vision}. The commonly used methods in ASSA are neural networks (NNs), Hidden Markov Models (HMMs), and Support Vector Machines (SVMs)~\cite{lam2022machine}. Recent research has shown that while SVMs and HMMs can still be effective for certain video classification tasks, their performance often lags behind more advanced deep learning methods, such as 3D CNNs or transformers, especially for large-scale and complex datasets~\cite{liu2025vision}. HMMs~\cite{peng2019automatic, gorantla2019surgical} can capture the temporal dependencies of the actions and can be trained using annotated surgical videos or physiological signals. In addition, it can capture the underlying patterns of the data and make probabilistic predictions. HMMs can’t learn a huge dataset and function as well as NNs. Nevertheless, these approaches can learn and perform acceptably using less data in some applications, such as ASSA~\cite{lam2022machine}. However, tuning the hyper-parameters of the HMMs is not straightforward~\cite{ahmidi2015automated,sun2017smart}. SVMs can handle non-linear decision boundaries through the use of kernel functions. However, SVMs are sensitive to the choice of kernel function and parameters. The computational complexity of SVM procedures is light, while it is an arduous process for multi-label and massive  datasets~\cite{ershad2019automatic}. According to the state-of-the-art research~\cite{lam2022machine}, the achievable performance accuracy of the SVMs for video classification is yet not satisfactory. In addition, CNNs can learn to estimate the locations of surgeons' tools from surgical videos, such as Temporal Segment Networks (TSN) and Faster Region-based CNN (Faster-RCNN)~\cite{wang2016temporal, ren2015faster}. CNNs have also been trained to provide interpretable, domain-specific feedback to trainees, such as highlighting suboptimal tool trajectories, pinpointing spatial movement errors via activation maps, or assigning objective skill scores~\cite{kiyasseh2023multi}. Although deep learning-based approaches can perform such assessments accurately~\cite{zhang2020automatic,funke2019video}, learning well requires very rich datasets and highly intricate computations.

While most ASSA methods primarily analyze robot kinematic or tool motion data, acquiring such data necessitates access to either a robotic surgical system or specialized tracking systems, which makes it complex and costly in practice~\cite{chen2019objective}. In contrast, video data can be effortlessly obtained in both traditional and robot-assisted minimally invasive surgery, primarily because modern operating rooms are now routinely equipped with high-definition endoscopic and external cameras that facilitate seamless data collection~\cite{rueckert2024methods, harari2024deep}. However, video data is high-dimensional and much more intricate than sequences of a few motion variables, which has also motivated automated preprocessing pipelines that shorten raw operative recordings by localising the relevant anatomy and instruments and discarding inactive frames before any further analysis~\cite{hatam2024aipowered}. The recorded video can be used to obtain kinematic data by tracking surgical tools and assess surgical skill by analyzing this data without the requirement of having motion tracking markers~\cite{bouget2017vision}. Such kinematic data can also be obtained using convolutional neural networks (CNNs). In~\cite{du2018articulated}, the authors propose a deep neural network for articulated multi-instrument 2-D pose estimation, which is trained on detailed annotations of endoscopic and microscopic data sets. The overall scheme can effectively localize instrument joints and also estimate the articulation model. The same perception requirement arises in ophthalmic microsurgery, where dedicated pipelines have been developed for the capsulorhexis phase: an expert-annotated framework for detecting the pupil and the cystotome~\cite{ahmadi2021arasfarabi}, and Siamese-network trackers that follow the instrument across frames~\cite{lafouti2022surgical}. In another approach, video data can be converted to time series for ASSA. In~\cite{zia2018video}, the authors use entropy-based features, approximate entropy and cross-approximate entropy, to find the difference in the predictability of motions. This method is based on the fact that an expert surgeon will have more predictable hand motion while a beginner will exhibit erratic and irregular patterns.

The two above methods discussed require intermediate steps, which brings in complexity. Furthermore, in~\cite{funke2019video}, both spatial and temporal information are captured using CNNs for surgical skill assessment. Thus, new studies have been focused on methods that exclusively rely on using surgical videos. In video-based action recognition and skill assessment, some strategies~\cite{lin2021efficient, kitaguchi2021development} demonstrate that utilization of optical flows of videos with RGB videos simultaneously causes further accuracy in classification. For this purpose,~\cite{funke2019video} has rendered a two-streamed CNN, which contains two parallel networks and gets two kinds of inputs: RGB frame sequences and optical flow feature maps, although satisfactory accuracy is obtained, the computations are very intricate. Performed experiments of such networks in surgical skill assessment demonstrate that accuracy has increased compared to that of the trained model with only RGB frames~\cite{li2019manipulation}. Computing optical flows of videos compels us to consume additional time. To overcome this issue,~\cite{piergiovanni2019representation} introduces a unified structure that uses optical flow features, although employing this structure in action recognition has been performed with better precision than two-streamed CNNs. Since this network is very promising, it has been applied in our surgical skill assessment framework. In another study~\cite{li2021real}, a real-time Gaussian process-based approach has been used for real-time ASSA using kinematic data.

Beyond this optical-flow lineage, video-based SSA has been pursued from several complementary directions by different groups. One line of work performs frame-level feature extraction with temporal aggregation, e.g., Temporal Segment Networks and related architectures that pool clip-level evidence over time~\cite{wang2016temporal}. A second direction couples convolutional encoders with recurrent or temporal-convolutional models to capture the sequential structure of a procedure and, in some cases, to localise the gestures that drive a skill rating. A third, increasingly influential direction targets interpretability and actionable feedback rather than a single score, e.g., multi-task models that jointly predict skill and generate trainee-facing feedback~\cite{kiyasseh2023multi}. A fourth body of work emphasises objective, contactless performance metrics and hand/instrument motion recovered directly from video~\cite{soleymani2024hands, zia2022objective}, while recent surveys document the rapid diversification of these methods~\cite{lam2022machine, liu2025vision}. Importantly, these efforts span surgical domains well beyond robotic bench-top tasks: video-based skill assessment has been studied extensively in laparoscopic and, more recently, in ophthalmic surgery; for instance, the capsulorhexis phase of cataract surgery, for which large multi-centre resources such as the Cataract-LMM benchmark now provide objective surgical skill scores across procedures using a multi-indicator rubric adapted from validated clinical standards~\cite{2026cataract}. In this domain the perception stack that precedes assessment has itself been the subject of dedicated work, for example pixel-level delineation of the pupil, the capsule and the instrument during capsulorhexis~\cite{gandomi2023deep}, and the wider role of image processing and machine vision in surgery and in surgical training has been reviewed in~\cite{ahmadi2023image}. A recurring theme across these studies is that generalisation across procedures, centres, and acquisition conditions remains an open challenge~\cite{kim2025generalizability}. Our work is positioned within this landscape: rather than competing on raw model capacity, we focus on the limited-data, computationally constrained regime, and we adopt the representation-flow backbone introduced above for its motion-awareness without the cost of explicit optical-flow precomputation.

Recent progress in video understanding has also been driven by transformer-based architectures (video vision transformers) that model long-range temporal dependencies via self-attention. These models often achieve strong performance when pretrained on large-scale video datasets and then fine-tuned to downstream tasks. However, their effective training typically requires substantial compute and careful regularization, which can be challenging in small-data surgical benchmarks.

To address the aforementioned limitations, we propose RF-DUAGP-Tree, a novel framework for surgical skill assessment that combines Representation Flow~\cite{piergiovanni2019representation}-based spatio-temporal feature extraction (RF) with a dynamics- and uncertainty-aware Gaussian Process Tree classifier (DUAGP-Tree) which is the enhanced version of GP-Tree~\cite{achituve2021gp}. The proposed framework introduces clip-level latent motion dynamics descriptors, including motion trend, acceleration, and jerk, which are jointly utilized as discriminative features and uncertainty indicators. Furthermore, semantic compositional uncertainty kernels are developed to explicitly model the heterogeneous semantics of surgical motion representations and improve robustness to temporal fluctuations and abnormal transitions. The proposed method is evaluated on the JIGSAWS~\cite{gao2014jhu} and the Cataract-LMM (hereafter referred to as the Capsulorhexis dataset) benchmarks~\cite{2026cataract}, demonstrating competitive performance in limited-data settings. We report the complexity and inference cost of the proposed classifier, while avoiding direct wall-clock superiority claims over published deep-learning baselines that were not rerun on our hardware.

The remainder of the paper is structured as follows. In Section~\ref{background}, an overview of the background and preliminaries is presented. Next, in Section~\ref{proposed method}, we discuss the proposed pipeline for SSA. Section~\ref{Results} provides the results of the proposed method compared to the existing ones, and Section~\ref{Discussion} discusses their implications and the limitations of the study. Finally, the concluding remarks are given in Section~\ref{Conclusion}.

\section{Background and Preliminaries}
\label{background}
\subsection{JIGSAWS}
The JHU-ISI Gesture and Skill Assessment Working Set (JIGSAWS)~\cite{gao2014jhu} is a publicly available surgical dataset created by researchers at Johns Hopkins University and the University of Southern California. The dataset contains recordings of surgical tasks performed by both experienced and novice surgeons using the da Vinci Surgical System. The JIGSAWS dataset includes videos of three surgical tasks: needle passing, suturing, and knot tying. The dataset also includes kinematic and tool usage data, such as the position and orientation of the surgical instruments, the amount of force applied by the surgeon, and the speed of the movements. JIGSAWS also contains annotations for surgical gestures and skill, which have been rendered manually, based on an experimental setup that comprises two standard cross-validations and a C++/Matlab toolkit to analyze surgical gestures employing hidden Markov models and utilizing linear dynamical systems. A summary of skill levels for different procedures of JIGSAWS is summarized in Table~\ref{table0}. As observed from the table the videos do not have the same number of frames and the number of videos for each task is limited.

\begin{table}
 \caption{The summary of the JIGSAWS dataset}\label{table0}
  \centering
  \begin{tabular}{llll}
    \toprule
    %\multicolumn{4}{c}{Part}                   \\
    \cmidrule(r){1-4}
    Label     & Suturing     & Knot-Tying & Needle-Passing \\
    \midrule
    Novice & 38 & 32 & 22\\
    Intermediate & 20 & 20  & 16 \\
    Expert & 20  & 20 & 18 \\ 
    \midrule
    Total Videos & 78 & 72 & 56\\
    Video Frames & 1794-9026 & 1014-2727 & 1789-4776\\
    \bottomrule
  \end{tabular}
  \caption*{Counts are given in videos; because both capture views of each trial are used, the corresponding trial counts are 39, 36 and 28 for Suturing, Knot-Tying and Needle-Passing. Both views of a trial are always assigned to the same cross-validation fold.}
\end{table}

\subsection{Capsulorhexis Dataset (Cataract-LMM)}
\label{CapsDS}
The Capsulorhexis dataset utilized in this study is derived from the Skill Assessment subset of the Cataract-LMM benchmark~\cite{2026cataract}. This specific subset comprises video clips focused exclusively on the capsulorhexis phase of phacoemulsification surgery. To support objective competency assessment, each clip is annotated with quantitative skill scores based on a video-based rubric developed by a consensus panel of consultant ophthalmic surgeons and medical education experts. The evaluation framework adapts six core performance indicators from the validated GRASIS and ICO-OSCAR standards: Instrument Handling, Tissue Handling, Microscope Use, Commencement of Flap, Circular Completion, and Motion. Each indicator is assessed on a 5-point scale, culminating in a continuous overall proficiency score for each procedure. For the binary classification experiments in this study, the continuous overall proficiency score was dichotomized using a fixed threshold of 3.8: scores below 3.8 were assigned to the Novice class, whereas scores greater than or equal to 3.8 were assigned to the Expert class. This cutoff was fixed before the repeated train/test evaluations and was applied unchanged across all three training runs and their associated testing rounds; it was not tuned separately for individual evaluations.

\subsection{Representation Flow}
\label{RFlow}
Representation Flow (RF)~\cite{piergiovanni2019representation} is the front end of the proposed framework and is used here exclusively as a motion-aware feature extractor. Because its parameters are fine-tuned in this work and its outputs define the descriptors $f$, $v$, $a$ and $j$, we summarise it here in a self-contained form.\\[5pt]
\textbf{Motivation.} Conventional two-stream action-recognition pipelines pre-compute a dense optical-flow field on the RGB frames using an iterative TV-L1 solver, store it, and process it with a second network. This is accurate but expensive: the flow computation is an offline pass over every frame pair and is not adapted to the recognition task. Representation Flow instead implements the TV-L1 iteration as a differentiable layer inside a single 3D convolutional network, and applies it not to pixels but to the network's own intermediate feature maps. Motion is therefore represented as the apparent displacement field of learned feature channels between adjacent temporal positions, it is computed within the forward pass, and the parameters of the flow computation are themselves trainable.\\[5pt]
\textbf{Architecture.} The backbone is a pretrained 3D CNN consisting of a stem (\texttt{conv3D}$\rightarrow$\texttt{BatchNorm3D}$\rightarrow$\texttt{maxpool3D}), two stages of 3D residual blocks, the flow layer, two further stages of 3D residual blocks, and a final pre-classification 3D convolution (Fig.~\ref{TS}). Two intermediate representations are used in this work and are referred to throughout as the two \emph{feature streams}: the output of the flow layer, $\Phi_{\mathrm{flow}}(c)\in\mathbb{R}^{512\times73}$, which encodes motion explicitly; and the output of the pre-classification layer, $\Phi_{\mathrm{pre}}(c)\in\mathbb{R}^{2048\times73}$, which encodes higher-level semantics. Here $c$ is a 300-frame input clip, the second axis (73) indexes the temporal positions remaining after the network's temporal striding, and the first axis indexes feature channels.\\[5pt]
\textbf{The flow layer.} Given two feature maps $F_t$ and $F_{t+1}$ at adjacent temporal positions, the layer estimates a displacement field $\bm{u}=(u_x,u_y)$ by unrolling a fixed number of iterations of the TV-L1 problem
\[
\min_{\bm{u}}\ \int \big(|\nabla u_x| + |\nabla u_y|\big)\,\mathrm{d}\Omega \;+\; \lambda \int \big|\rho(\bm{u})\big|\,\mathrm{d}\Omega,
\qquad
\rho(\bm{u}) = \nabla F_{t+1}\cdot(\bm{u}-\bm{u}_0) + F_{t+1} - F_t,
\]
where $\rho$ is the linearised feature-constancy residual. Each unrolled iteration performs (i)~a dual ascent step on the dual variables $\bm{p}$ with step size $\tau$, (ii)~a proximal (soft-thresholding) step on an auxiliary field whose threshold is set by $\lambda\theta$, and (iii)~a coupling step controlled by $\theta$. Spatial gradients and the divergence are computed by the fixed convolution kernels $w_x$ and $w_y$. The parameters have the following roles:
\begin{itemize}
  \item $\lambda$ --- the weight of the data (feature-constancy) term relative to the total-variation smoothness term. Larger $\lambda$ makes the estimated field follow the observed feature differences more closely, giving sharper but noisier motion; smaller $\lambda$ yields a smoother, more strongly regularised field.
  \item $\theta$ --- the tightness parameter of the convex relaxation that decouples the data term from the TV term. It controls how tightly the auxiliary field is tied to $\bm{u}$ and, with $\lambda$, sets the width of the soft-threshold; small $\theta$ gives a tighter but more slowly converging solution.
  \item $\tau$ --- the step size of the dual update on $\bm{p}$, governing the stability and convergence rate of the unrolled iterations; for the standard four-neighbour discretisation, stability requires $\tau\le1/4$.
  \item $w_x$, $w_y$ --- the convolution kernels implementing the horizontal and vertical spatial-derivative operators (and, transposed, the divergence). They may in principle be learned; following~\cite{piergiovanni2019representation}, they are held fixed at their Sobel-like initial values, which we found to be more stable and more accurate.
\end{itemize}
All five parameters are differentiable. In this work $\lambda$, $\theta$ and $\tau$ are updated during the short adaptation stage described in Section~\ref{FAR}, while $w_x$ and $w_y$ are kept at their initial values; the number of convolutional blocks of the backbone that are fine-tuned is given in Table~\ref{tablemodelcapcity}.

\subsection{Gaussian Processes Classification}
\label{GPC}
Gaussian process (GP) is a non-parametric Bayesian approach for active learning and optimizing unknown functions for both regression and classification. Contrary to NN methods, GP is less dependent on the massive training data sets for generalization. This will help the development of an ASSA pipeline with limited training data. Since GP contains few training parameters, it is also computationally efficient. Thus, it does not add computational complexity when being used along with NNs. In what follows, we will discuss the preliminaries for GP classification.

Suppose having a dataset as: \{$(\boldsymbol{x}_1,y_1), \cdots,(\boldsymbol{x}_N,y_N)$\} where $N$ is the number of data with $C$ different labels, such that: $y_i \in \{1,..., C\}$. 
For classification purpose it is desired to find latent function $F \in \mathcal{R}^{C \times N}$ such that $p(y_i |((f^1(\boldsymbol{x}_i),...,f^C(\boldsymbol{x}_i))))$ estimates the corresponding label of $\boldsymbol{x}_i$. For multi-class GPC where $C \geq 3$, $F$ can be written in the following form:

\begin{equation}
F=
%\[
\left (
\begin{array}{ccc}
\begin{array}{l}
f^1(\boldsymbol{x}_1)\\
f^2(\boldsymbol{x}_1)
\end{array}
& \cdots & 
\begin{array}{l}
f^1(\boldsymbol{x}_N)\\
f^2(\boldsymbol{x}_N)
\end{array} \\
\vdots & \ddots & \vdots\\
\begin{array}{l}
f^{c-1}(\boldsymbol{x}_1)\\
f^{c}(\boldsymbol{x}_1) 
\end{array} &
\cdots & 
\begin{array}{l}
f^{c-1}(\boldsymbol{x}_N)\\
f^{c}(\boldsymbol{x}_N)  \\
\end{array} 
\end{array}
\right )
%\]            
\end{equation}

In multi-class GPC, contrary to Gaussian process regression and binary classification, more than one multivariate Gaussian distribution should be considered, such that for each label a multivariate Gaussian distribution should be defined as $\mathbf{f}_k \sim GP(0, K_k)$, while $\mathbf{f}_k = \{f^k(\boldsymbol{x}_1),...,f^k(\boldsymbol{x}_N)\}$, and $K_k$ is the relevant kernel function. Generally, for estimating labels, Eq.~\eqref{eq2} can be used:

\begin{equation}
p(\mathbf{y}|F) = \prod\limits_{i=1}^{N}p\left(y_i |\left(f^1\left(\boldsymbol{x}_i\right),...,f^C\left(\boldsymbol{x}_i\right)\right)\right)
\label{eq2}
\end{equation}

Such a classification is computationally intense. Kim and Ghahramani~\cite{kim2006bayesian} introduced a method to reduce the complexity to $O(N^3)$. This complexity can be reduced to $O(MN^2)$ by Sparse Gaussian processes using pseudo-inputs~\cite{snelson2005sparse}. In several previous works, the right-hand side of Eq.~\eqref{eq2} is replaced with a soft-max likelihood \cite{kim2006bayesian,williams1998bayesian} for multi-classification purposes, however, this is not scalable with the number of data instances, due to computational cost. As a result, \cite{hernandez2011robust} has presented a scalable multi-class GPC using the robust-max likelihood. Authors in \cite{villacampa2017scalable} have rendered a scalable multi-class GPC using expectation propagation by the Heaviside likelihood. However, these methods are still slow and computationally complicated.
To tackle complications, other solutions based on probabilistic data augmentation are introduced. For example, in~\cite{linderman2015dependent}, authors introduced a sequential method named Multinomial Stick Breaking structure using Pólya--Gamma augmentation for further simplification of the computational complexity.

\begin{figure*}
\centering
\subcaptionbox[]{\label{fig1a}}{\includegraphics[width=10cm]{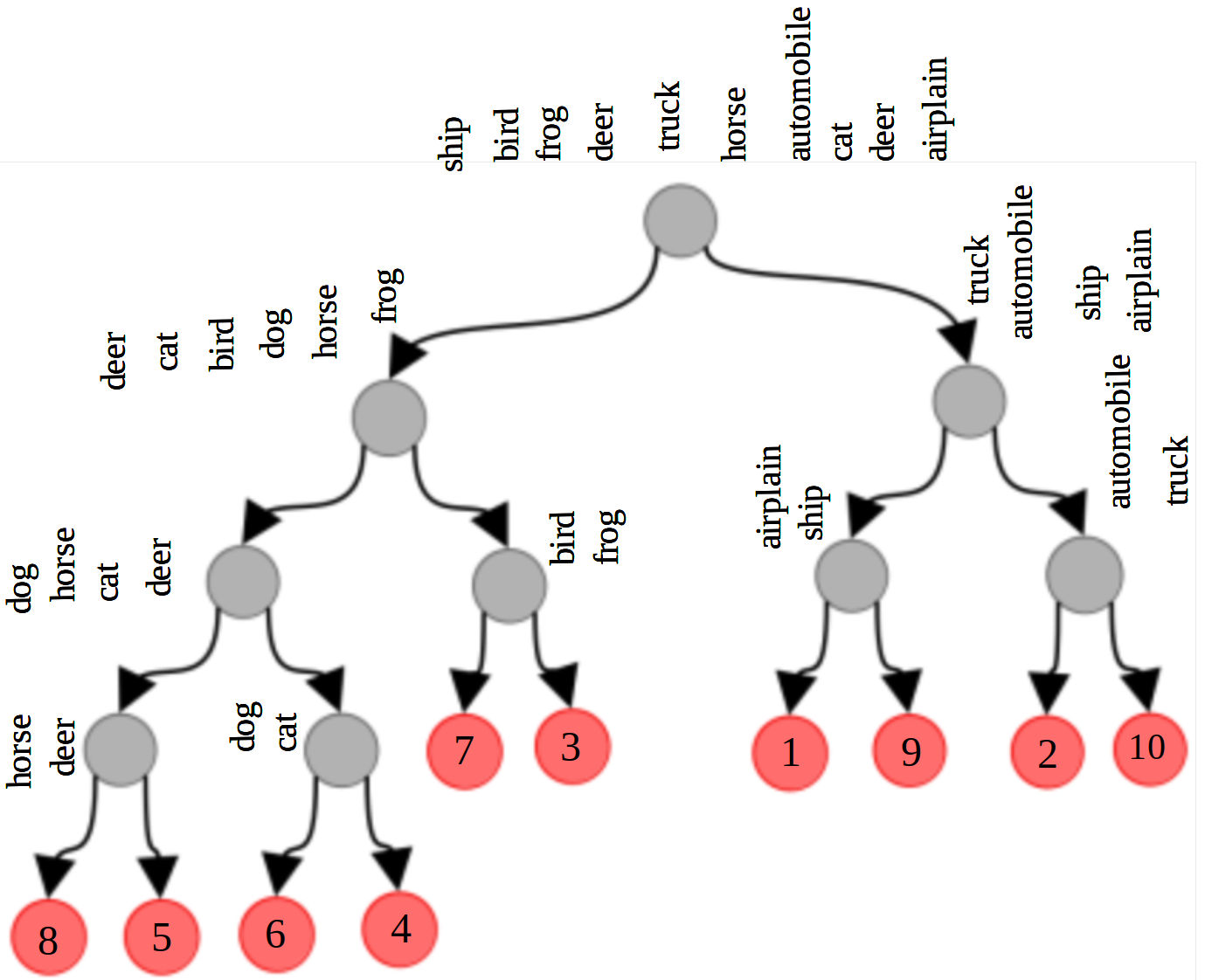}}%
\hfill
\subcaptionbox[]{\label{fig1b}}{\includegraphics[width=4cm]{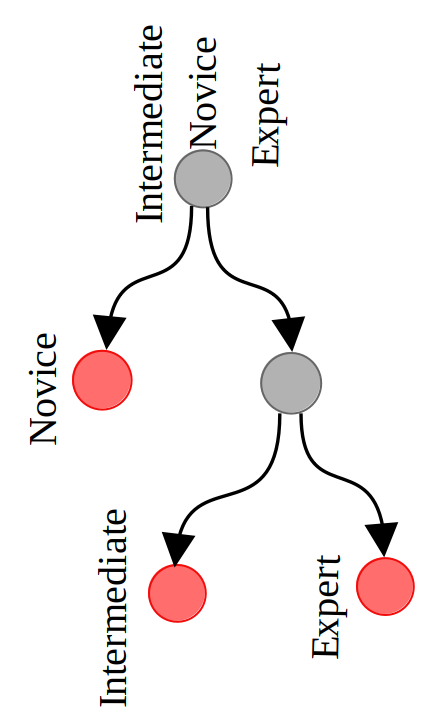}}%

\caption{(a) The structure of GP-Tree based on CIFAR-10~\cite{achituve2021gp},
(b) The structure of GP-Tree based on JIGSAWS.}
\end{figure*}

In another study, the authors introduced a hierarchy-based method to reduce the complexity of multi-class GPC, named GP-Tree for image classification using deep neural networks~\cite{achituve2021gp}. It is a tree-based model in which individual nodes leverage a sparse model of Gaussian processes by the Pólya--Gamma augmentation to perform binary classification tasks. The structure of the tree is designed based on the number of labels. As an example, Figure~\ref{fig1a} illustrates the tree model for the CIFAR-10 data set. 
As observed from this figure, GP-Tree generates a balanced tree that is divided by the semantic meaning of the classes. For example for CIFAR-10 data, motorized vehicles are on the right subtree of the root node while animals are on the left one. This semantic partition is pronounced at all tree levels. The simpler tree structure used for the three-class JIGSAWS dataset in this study is illustrated in Figure~\ref{fig1b}. One important aspect of this method is the architecture of the tree, where the authors discuss a method based on $k$-means++ to assign labels to the nodes, which was previously introduced in~\cite{arthur2006k}. In this method, the latent functions of the GP, $\mathbf{f}_v$ for internal nodes $v$, can be obtained using:

\begin{equation}
\mathbf{f}_v \sim GP(\mathbf{m}_v , \mathbf{K}_v)
\end{equation}

in which $\mathbf{m}_v$ and $\mathbf{K}_v$ are the mean function and covariance function, respectively. 
Each GP can learn whether any input should go to the left child node or the right one in an unsupervised way. Two label arrays are allocated for each internal node (right $= 0$ and left $= 1$). Leaves are expressive of major labels, and for all the inputs from any class $c$, there is only a unique path from the root to the leaf of that class label ($P_c$), which results in estimating a label as follows:
\begin{equation}
p(y=c|\mathbf{F}) = \prod_{v \in P_c} \sigma(f_v)^{y_v}(1-\sigma(f_v))^{1-y_v}
\end{equation}

where $\mathbf{F}$ is a set of all the latent functions of the GP models in the tree and $\sigma$ is the logistic likelihood function. The GP model for each node uses Pólya--Gamma Data Augmentation~\cite{wenzel2019efficient} in which $X=(\boldsymbol{x}_1,...,\boldsymbol{x}_N) \in \mathbb{R}^{K \times N}$ are inputs of a GP and $\mathbf {y} = (y_1,...,y_N) \in \{0, 1\}^N$ are the labels which are outputs. Thus, the model can be defined as:

\begin{equation}
p(\mathbf{y,w,f,u},X) = p(\mathbf{y}|\mathbf{f,w})p(\mathbf{w})p(\mathbf{f}|\mathbf{u},X)p(\mathbf{u})
\label{eqq}
\end{equation}

In Eq.~\eqref{eqq}, $\mathbf w$ contains Pólya-Gamma variables, $\mathbf f$ is a vector of latent decision functions of $X$, and $\mathbf u$ indicates a vector of latent decision functions of inducing points.  Using the above equation, the prediction of the labels for the test data can be represented as:
\[
  \begin{split}
   p(f_*|\boldsymbol{x}_*, \bar{\mathbf{X}}, \bar{\mathbf{y}}) \approx \int p(f_* | \mathbf{u},\boldsymbol{x}_*)q(\mathbf{u}) d\mathbf{f}\\
   = N(f_* | \mu_*,\scriptstyle \sum_*),\\
  \mu_* = \mathbf{k}_{m*}^{\top}\mathbf{K}_{mm}^{-1}\tilde{\bm{\mu}},\\
  \qquad
  \bm{\Sigma}_* = k_{**} - \mathbf{k}_{m*}^{\top}\big(\bm{\Sigma}\mathbf{K}_{mm}^{-1} -       \mathbf{I}\big)\mathbf{k}_{m*},
  \end{split}
\]

where $\bar{\mathbf{X}}$ denotes the inducing points, $\bar{\mathbf{y}}$ are the labels~\cite{wenzel2019efficient}, and $m$ is the number of inducing points, while $\tilde{\bm{\mu}}$ and $\bm{\Sigma}$ are the mean and the covariance of the variational distribution over the inducing values.
In order to do inference in each node, variational inference has been employed. $q$ is the variational distribution. Additionally, $\mathbf{ K}_{mm}$ is the kernel for $m$ inducing points, while $*$ is used to represent test data instances. Notably, the computational complexity of any binary decision GP is $O(m^3)$. As a result, the computational cost of GP-Tree is about $O((C-1)m^3)$, which is substantially less than the general GPC.

\section{Proposed Method}
\label{proposed method}
In this section, we elaborate on the proposed pipeline for ASSA using video data. The proposed method is a 3D CNN with an inherent representation flow algorithm (Section~\ref{RFlow}), originally introduced in~\cite{piergiovanni2019representation}, coupled with DUAGP-Tree (the dynamics- and uncertainty-aware GP-Tree), referred to as RF-DUAGP-Tree. As discussed earlier, DUAGP-Tree is based on the GP-Tree method.  DUAGP-Tree is robust to abrupt temporal fluctuations and abnormal transitions, while utilizing latent motion dynamics, especially as discriminative features introduced in this study. In other words, latent motion dynamics play two crucial and contrasting roles in the method: 1. dynamic features, and 2. uncertainty.

\begin{figure*} [!ht]
\centering
\includegraphics[width=\linewidth]{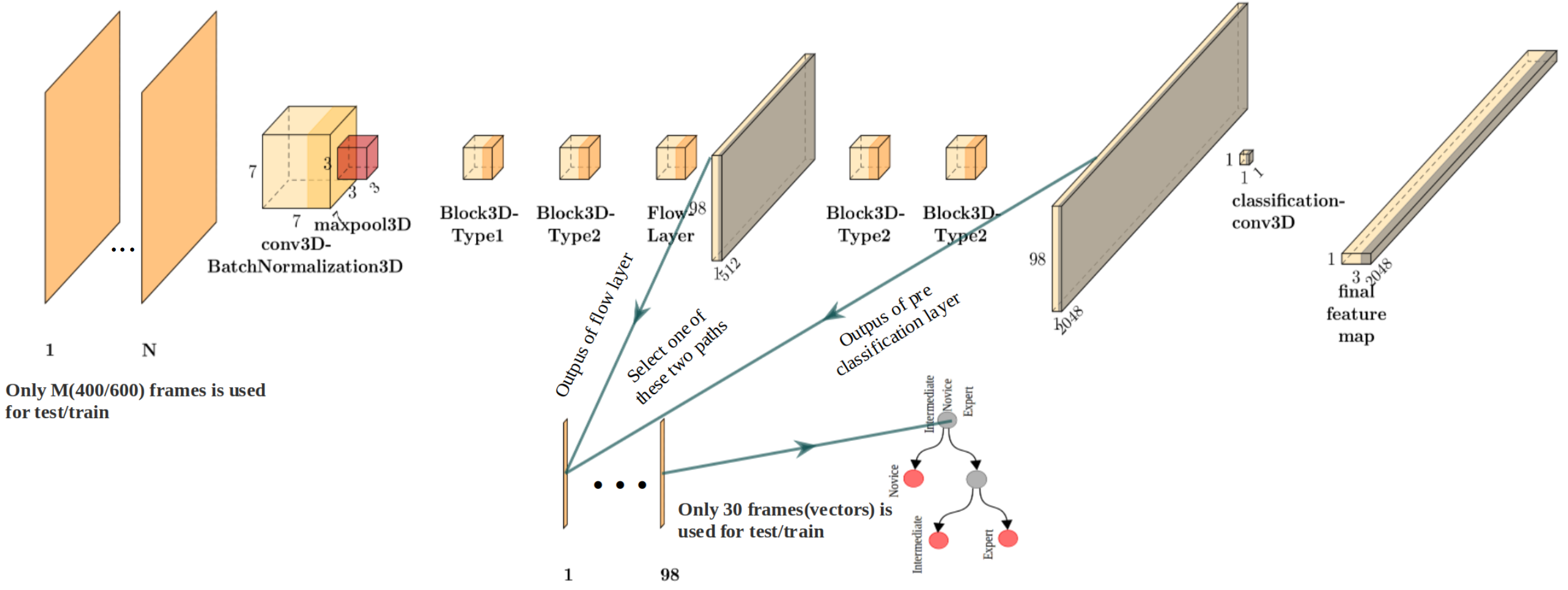}
\caption{The structure of the RF-based surgical skill assessment strategies. A drawn instance for a knot-tying video. The structure shows different combinations of the RF combined with different classification methods. The two feature streams used throughout the paper are the output of the flow layer ($512$ channels) and the output of the pre-classification layer ($2048$ channels), both with $73$ temporal positions per 300-frame clip.}
  \label{TS}
\end{figure*}

Compared with the deep baselines published for the Capsulorhexis benchmark~\cite{2026cataract}, the proposed framework demonstrates superior data efficiency by achieving competitive performance with a substantially reduced training set (Section~\ref{modelcapcity}). For example, on the Capsulorhexis dataset only 30 training videos are required, whereas those baselines use approximately 140 videos for training. A schematic of the proposed pipeline is depicted in Figure~\ref{TS}.

In contrast to the GP-Tree framework introduced in~\cite{achituve2021gp}, the proposed DUAGP-Tree eliminates the Deep Kernel Learning (DKL) stage. Instead, classification is performed using engineered representations derived from the feature maps generated by the Representation Flow (RF) network. As these representations already encode rich spatiotemporal motion information, an additional deep kernel transformation becomes unnecessary. This design not only simplifies the overall architecture but also reduces computational complexity while preserving discriminative capability.

The DUAGP-Tree classifier serves as the primary decision-making component of the proposed framework. The Representation Flow (RF) network is utilized exclusively for feature extraction and is initialized with pretrained weights. To account for task-specific characteristics, only a lightweight fine-tuning stage is performed on the RF network before extracting feature representations. These representations are then provided to the DUAGP-Tree classifier, thereby decoupling feature extraction from probabilistic classification and reducing the overall training burden.

The notation used throughout the paper is collected in Table~\ref{tabNotation}, and the shape of every tensor along the data path is given in Table~\ref{tabShapes}.

\begin{table}
\centering
\caption{Notations}
\label{tabNotation}
\begin{tabular}{@{}p{0.16\textwidth} p{0.49\textwidth} p{0.26\textwidth}@{}}
%\toprule
%\textbf{Symbol} & \textbf{Meaning} & \textbf{Shape / range} \\
%\midrule
%\endfirsthead
\toprule
\textbf{Symbol} & \textbf{Meaning} & \textbf{Shape / range} \\
\midrule
%\endhead
$V$, $F$ & a surgical video and its number of frames & --- \\
$c_t$, $n$ & the $t$-th 300-frame clip; number of clips in a video & $t=1,\dots,n$ \\
$\Delta_c$ & temporal interval of one clip & 300 frames ($\approx10$\,s at 30\,fps) \\
$s$ & feature stream index, $s\in\{\mathrm{flow},\mathrm{pre}\}$ & --- \\
$D_s$ & channel dimension of stream $s$ & $512$ (flow), $2048$ (pre) \\
$T$ & temporal positions per clip after striding & $73$ \\
$f^{(s)}$ & latent spatio-temporal representation & $n\times D_s\times T$ \\
$v$ & motion trend (first backward difference over clips) & $(n{-}1)\times D_s\times T$ \\
$a$ & acceleration (second backward difference) & $(n{-}2)\times D_s\times T$ \\
$j$ & jerk (third backward difference) & $(n{-}3)\times D_s\times T$ \\
$X$ & video-level feature vector $[f,v,a,j]$ presented to the classifier & $D_{\mathrm{in}}$; $27$ for Capsulorhexis \\
$\lambda,\theta,\tau$ & flow-layer data weight, relaxation tightness, dual step size & scalars \\
$w_x, w_y$ & fixed spatial-derivative kernels of the flow layer & $3\times3$ \\
$C$, $y_i$ & number of skill classes; label of sample $i$ & $C=3$ (JIGSAWS), $2$ (Caps.) \\
$N$ & number of training samples (videos) in the current fold & $46$--$68$ (JIGSAWS); $30$ (Caps.) \\
$d$ & feature-dimension index & $d=1,\dots,D$ \\
$\mathbf{f}_v$, $P_c$ & latent GP function at tree node $v$; root-to-leaf path of class $c$ & --- \\
$m$, $\bar{\mathbf{X}}$ & number and locations of inducing points & $m\in\{2,8,15\}$ \\
$\mathbf{K}_{mm}$, $\mathbf{k}_{m*}$ & inducing-point kernel matrix; test-to-inducing kernel vector & $m\times m$; $m\times1$ \\
$\tilde{\bm{\mu}}$, $\mathbf{S}$ & variational mean and covariance of the inducing values & $m\times1$; $m\times m$ \\
$\bm{\Sigma}_*$ & predictive covariance & --- \\
$\bm{\Sigma}_x$ & input-uncertainty covariance (diagonal) & $D\times D$ \\
$\sum_t$ & summation over the clip index (range stated at each use) & --- \\
$H(\cdot)$, $\epsilon$ & entropy operator; numerical floor inside the logarithm & scalar; $10^{-8}$ \\
$\sigma^2(\cdot)_t$ & variance taken over the clip index & scalar per dimension \\
$\ell_d$ & ARD length-scale of feature dimension $d$ & scalar \\
$k_f,k_v,k_a,k_j$ & semantic sub-kernels of the four feature blocks & Gram matrices \\
$k_e$, $k_t$ & embedding and temporal components of $k_f$ (JIGSAWS) & Gram matrices \\
$K_{\mathrm{flow}}, K_{\mathrm{end}}$ & stream-specific kernels of Eq.~\eqref{kfe} & Gram matrices \\
$\alpha,\beta,\gamma,\delta,a,b$ & kernel composition weights & scalars \\
$\lambda_v,\lambda_a,\lambda_j$ & uncertainty weights, fixed to $1$ & scalars \\
$L_d^{(\cdot)}$ & uncertainty-derived adaptive length-scale (UA1/UA2/EA) & scalar per dimension \\
$\odot$ Eq.~\eqref{kj} & elementwise (Hadamard) product of Gram matrices & --- \\
$\top$ & matrix/vector transpose & --- \\
\bottomrule
\end{tabular}
\end{table}

\begin{table}
\centering
\caption{Tensor shapes along the data path}
\label{tabShapes}
\begin{tabular}{@{}p{0.30\textwidth} p{0.32\textwidth} p{0.32\textwidth}@{}}
\toprule
\textbf{Stage} & \textbf{Shape} & \textbf{Note} \\
\midrule
Input video & $3\times F\times720{\times}480$ or $640{\times}480$ & resized to $112\times112$ \\
Clip $c_t$ & $3\times300\times112\times112$ & $n=\lfloor F/300\rfloor$ clips \\
Flow-layer tap & $512\times73$ per clip & motion representation \\
Pre-classification tap & $2048\times73$ per clip & semantic representation \\
$f^{(s)}$ & $n\times D_s\times73$ & $D_s\in\{512,2048\}$ \\
$v$, $a$, $j$ & $(n{-}1)$, $(n{-}2)$, $(n{-}3)\times D_s\times73$ & backward differences over $t$ \\
Capsulorhexis reduction & $5+8+8+6 = 27$ per video & $v,a,j$: mean, std, energy, entropy per stream (mean omitted for $j$); $f$: clip count $+$ mean, std \\
JIGSAWS reduction & $f_d$: $n\times D_s$; $f_t$: $n\times73$; $v,a,j$: mean-pooled & $k_f = k_e + k_t$, Eq.~\eqref{kfkekt} \\
Classifier input & $D_{\mathrm{in}}$ per video & one sample per video \\
\bottomrule
\end{tabular}
\end{table}

\subsection{DUAGP-Tree}
\label{DUAGPT}
The proposed DUAGP-Tree retains the key advantages of Gaussian Process (GP)-based models, including effective learning from limited training data and reduced computational requirements. As described previously, the classifier operates on representations derived from the feature maps generated by the Representation Flow (RF) network. These feature maps are extracted at the clip level, where each surgical video is partitioned into a sequence of consecutive clips. Clips are an internal device for handling videos of unequal length and are never treated as independent samples: the feature maps of all clips of a video are aggregated into a single video-level representation with a single label before classification, so that the video is the atomic unit of every train/test partition. To obtain a video-level representation, the feature maps corresponding to successive clips are aggregated and concatenated, resulting in a unified representation for each video.

To serve as inputs to the DUAGP-Tree classifier, the extracted feature maps are transformed into semantically meaningful feature representations. Each feature map encodes both spatial and temporal information and can be viewed as a sequence of representations obtained from consecutive video frames within a clip. Consequently, each normalized feature map is transformed into a semantic feature representation consisting of four complementary components:

\begin{center}
\begin{equation}
X = [f, v, a, j]
\end{equation}

\end{center}

where $f$ denotes the latent spatio-temporal feature representation extracted from the Representation Flow network. The component $v$ represents the motion trend which corresponds to the difference between two consecutive clips. This component preserves the direction of motion in the latent feature space and therefore encodes valuable information regarding the progression of surgical actions. The component $a$ denotes the acceleration, representing the second-order temporal variation of the latent features. This term captures motion instability and irregular transitions that are commonly associated with inconsistent surgical performance. For example, novice surgeons often exhibit less smooth and less controlled hand movements than experts, resulting in larger fluctuations in the latent feature trajectories. Finally, the jerk component $j$ is computed as the temporal variation of acceleration. The three components are defined as:
%\begin{center}
%\begin{equation}
\[
\begin{aligned}
  v_t &= f_t - f_{t-1}, && t=2,\dots,n,\\
  a_t &= v_t - v_{t-1} \;=\; f_t - 2f_{t-1} + f_{t-2}, && t=3,\dots,n,\\
  j_t &= a_t - a_{t-1}, && t=4,\dots,n.
\end{aligned}
\]
%\end{equation}
%\end{center}
The jerk term characterizes higher-order motion dynamics, including smoothness, stability, continuity, abrupt corrections, and tremor-like movements. As a result, $j$ provides additional discriminatory information for distinguishing different levels of surgical expertise.

Overall, the proposed derivative-based feature representation provides an approximation of the geometric structure of latent motion trajectories by modeling temporal derivatives on the underlying feature manifold. The resulting representation captures not only static spatio-temporal information but also the evolution of surgical motions through velocity, acceleration, and jerk components. The two instantiations are named and defined here, and the measurable dataset characteristics that motivate the choice between them are tabulated in Table~\ref{tabDataChar}:

\begin{itemize}
  \item \textbf{Variant-S (structured)}, used for JIGSAWS: the latent representation $f$ is retained as separate feature-distribution and temporal components combined through $k_f = k_e + k_t$ of Eq.~\eqref{kfkekt}; $v,a,j$ are temporally mean-pooled; the two RF streams are modelled by separate kernels and combined through the stream composition $k = a K_{\mathrm{flow}} + b K_{\mathrm{end}}$ of Eq.~\eqref{kfe}.
  \item \textbf{Variant-C (compact)}, used for Capsulorhexis: each block is reduced to summary statistics (27 features per video in total), and no stream-specific kernel decomposition is applied.
\end{itemize}

This dataset-aware design enables the Gaussian Process classifier to more effectively exploit the semantic and dynamic information embedded in surgical video representations.

\textbf{Capsulorhexis}

The dimensionality of the extracted feature maps is excessively high for efficient function inference within the DUAGP-Tree framework. Furthermore, the derivative-based representations ($v$, $a$, and $j$) may introduce additional noise due to the amplification of temporal fluctuations. To address these challenges, the feature maps are transformed into compact summary representations that preserve the most informative characteristics while substantially reducing dimensionality.

Specifically, each video-level feature representation $f$ is summarized using its mean, standard deviation, and the number of clips. Similarly, the derivative-based representations $v$, $a$, and $j$ are characterized using a set of descriptive statistics, including the mean, standard deviation, energy, and entropy. These summary features provide a compact representation of the underlying motion dynamics while maintaining the discriminative information required for skill assessment.

It should be noted that two distinct feature streams are extracted from the Representation Flow network. The first stream consists of features obtained from the representation flow layer, whereas the second stream is derived from the final layer of the network's last block. These feature streams capture complementary aspects of surgical motion and are investigated separately through the ablation studies. Experimental results indicate that the most effective configuration is obtained by jointly utilizing both feature streams.

This design is particularly advantageous for Gaussian Process learning, as compact yet informative representations improve computational efficiency while mitigating the risk of overfitting in limited-data settings.

The compound kernel formulations developed for the Capsulorhexis dataset are described in the following subsection.

\textbf{JIGSAWS}

In contrast to the Capsulorhexis dataset, the feature maps extracted from JIGSAWS contain rich discriminative information that should be preserved rather than aggressively summarized. Consequently, the primary feature representations $f$ are utilized directly within the proposed framework. Specifically, these representations are incorporated through separate embedding and temporal components using the compositional kernels introduced in Section~\ref{NIKernels}.

The derivative-based representations ($v$, $a$, and $j$) inherently encode temporal dynamics through their latent feature trajectories. Therefore, preserving their full temporal dimension is unnecessary. Instead, temporal mean pooling is applied to each representation, yielding compact descriptors that retain the essential dynamic characteristics while reducing dimensionality and computational complexity.

Similar to the Capsulorhexis setting, two complementary feature streams are considered: (i) features extracted from the representation flow layer and (ii) features obtained from the pre-classification layer of the network. These streams are incorporated through separate semantic compositional kernels, as described in Section~\ref{NIKernels}. The proposed kernels are designed to jointly exploit the complementary information contained in the latent feature representation $f$ and its derivative-based dynamic descriptors ($v$, $a$,$j$).

 Overall, the proposed semantic-aware compound kernels are specifically designed to capture the complementary semantic and dynamic information encoded in the latent feature representation and its temporal derivatives. These kernels form a key component of the DUAGP-Tree framework, enabling effective modeling of surgical motion patterns across different skill levels. In addition, the proposed uncertainty-aware kernels, presented in Section~\ref{NIKernels}, extend this formulation by explicitly incorporating uncertainty information derived from latent motion dynamics, thereby enhancing robustness to temporal variability and abnormal transitions.

To incorporate the second role of latent motion dynamics, namely uncertainty modeling, the covariance formulation of the proposed DUAGP-Tree is inspired by previously developed robust Gaussian Process methodologies. The underlying motivation is that temporal fluctuations, motion irregularities, and abrupt transitions observed in surgical videos can be interpreted as a source of input uncertainty and should therefore be explicitly accounted for during probabilistic inference.

Previous studies~\cite{mchutchon2011gaussian,johnson2019accounting} introduced robust Gaussian Process formulations capable of handling uncertainty and noise in the input space. In~\cite{mchutchon2011gaussian}, Gaussian Process models are learned directly from noisy inputs, whereas~\cite{johnson2019accounting} assumes that the inputs are first denoised before model construction. Inspired by these formulations, we extend the concept from input-noise modeling to uncertainty induced by latent motion dynamics.

Accordingly, we propose a robust sparse Gaussian Process classification framework based on Pólya--Gamma data augmentation. Following the covariance propagation principles established in the aforementioned studies, the predictive covariance $\bm{\Sigma}_*$ of the sparse predictive distribution given in Section~\ref{GPC} is redefined as:

\begin{center}
\[ \scalebox{1.25}{$\scriptstyle \sum_* = T_{**} + k_{**} - \mathbf k_{m^*}^T(\tilde{\mathbf \scriptstyle \sum}(\mathbf K^{-1}_{mm}) - \mathbf I)\mathbf k_{m^*}$} \]
\[ \scalebox{1}{$T_{ij} = T(\boldsymbol{x}_i, \boldsymbol{x}_j)=\partial _{u}(\boldsymbol{x}_i)(\scriptstyle \mathbf \sum_x)\textstyle \partial _{u}(\boldsymbol{x}_j)$} \]
\[ \scalebox{1}{$ \partial _{u}(\boldsymbol{x}_i) = \dfrac{\partial \mu_*}{\partial {\boldsymbol{x}_i}} $} \]
\end{center}

\setlength{\unitlength}{0.20mm}
where, $\mathbf \sum_x$ is the uncertainty variance, which is obtained by:

%\begin{equation}
\[
\bm{\Sigma}_x \;=\; \lambda_v \sum_{t=2}^{n}\sigma^2\!\big(H(\mathbf{v}_t\mathbf{v}_t^{\top})\big)
\;+\; \lambda_a \sum_{t=3}^{n}\sigma^2\!\big(H(\mathbf{a}_t\mathbf{a}_t^{\top})\big)
\;+\; \lambda_j \sum_{t=4}^{n}\sigma^2\!\big(H(\mathbf{j}_t\mathbf{j}_t^{\top})\big)
\;+\; \sigma^2 \mathbf{I},
\]
\label{s1}
%\end{equation}

or:

\[
\bm{\Sigma}_x \;= \lambda_j \sum_{t=4}^{n}\sigma^2\!\big(H(\mathbf{j}_t\mathbf{j}_t^{\top})\big)
\;+\; \sigma^2 \mathbf{I},
\]
\label{s2}

The entropy operator is computed using the entropy of the corresponding outer-product matrix,

\begin{center}
$ H(X)=-\sum_i p_i\log (p_i +\epsilon) $ 
\end{center}

in which $p_i = |z_i| / \sum_k |z_k|$ is the normalised magnitude distribution over the entries $z_i$ of the argument matrix, the normalisation is taken over the entries of that matrix, the logarithm is natural, and $\epsilon = 10^{-8}$ is a numerical floor. This provides a measure of the structural complexity and variability of latent motion dynamics. Consequently, highly irregular motion trajectories produce larger entropy values and contribute more strongly to the uncertainty covariance, whereas smooth and consistent surgical motions induce lower uncertainty levels.

Unless otherwise stated, the weighting coefficients $\lambda_v$, $\lambda_a$, and $\lambda_j$ are fixed to 1 in all experiments. Preliminary analyses indicated that the performance of the proposed framework is not sensitive to these coefficients; therefore, no additional hyperparameter optimization was performed. The uncertainty formulation is primarily intended to reflect relative variations in latent motion dynamics rather than to introduce additional trainable parameters.

Algorithm~\ref{alg1} summarizes the complete DUAGP-Tree framework.

\subsection{Newly Discovered Kernels: Semantic Compositional Uncertainty Kernels (SCUK)}
\label{NIKernels}

The proposed kernels are designed to jointly exploit the complementary information contained in the latent feature representation f and its derivative-based dynamic descriptors $v$, $a$, $j$. Since these components encode distinct semantic and dynamic characteristics of surgical motion, modeling them through a single homogeneous kernel may fail to fully capture their individual contributions. Therefore, semantic-aware compound kernels are introduced to explicitly account for the heterogeneous nature of the proposed feature representation.

More specifically, $f$, $v$, $a$, and $j$ represent complementary aspects of latent surgical motion, including appearance-aware spatio-temporal representations, motion trends, acceleration patterns, and higher-order motion variations, respectively. Consequently, separate kernel components are assigned to each feature group and subsequently combined into a unified compositional kernel.

\begin{equation}
k = \alpha k_f + \beta k_v + \gamma k_a + \delta k_j
\label{kx}
\end{equation}
\begin{equation}
k = \alpha k_f + \beta k_v + \gamma k_a 
\label{kav}
\end{equation}

\begin{equation}
k = \alpha k_f + \beta k_v + \gamma k_a + \delta \left( k_j \odot k_{entropy} \right)
\label{kj}
\end{equation}

The kernel formulations in Eqs.~\eqref{kx}--\eqref{kj} progressively incorporate different levels of latent motion information. While Eq.~(\ref{kx}) exploits the complete feature representation, including the jerk component, Eq.~(\ref{kav}) evaluates the contribution of lower-order motion dynamics by excluding jerk-related information. Finally, Eq.~(\ref{kj}) introduces entropy-aware modulation of higher-order dynamics, enabling the kernel to emphasize motion patterns associated with increased uncertainty and irregular temporal behavior. Here $\odot$ denotes the elementwise (Hadamard) product of the two Gram matrices, so that $k_j \odot k_{entropy}$ is a product kernel; $k_{entropy}$ is an ARD-RBF kernel evaluated on the entropy descriptors and contributes a single additional length-scale parameter on JIGSAWS and four on Capsulorhexis (Table~\ref{tablemodelcapcity}).

In addition to semantic decomposition, two complementary feature streams extracted from the Representation Flow network are jointly modeled. Specifically, flow-layer features capture explicit motion-related information, whereas pre-classification features provide higher-level semantic representations. To exploit both sources of information, the following stream-composition kernel is employed:

\begin{equation}
k = a K_{flow} + b K_{end}
\label{kfe}
\end{equation}

Each stream-specific kernel may itself correspond to any of the semantic compositions introduced above, allowing the proposed framework to simultaneously model feature semantics and multi-stream representations.

The stream-composition kernel in Eq.~(\ref{kfe}) is specifically introduced for the JIGSAWS dataset. Empirical analysis revealed that the flow-layer and pre-classification representations provide highly complementary information and exhibit substantially different embedding characteristics. Consequently, modeling these feature streams separately and subsequently combining them at the kernel level leads to a considerable improvement in predictive performance.

In contrast, for the Capsulorhexis dataset, the performance difference between single-stream and dual-stream representations was marginal: Strategy~1 (both streams) and Strategy~4 (one stream) both reach $0.892$ accuracy with a best AUC of $0.974$ (Table~\ref{table3}; see also Section~\ref{NDF}). Therefore, stream-specific kernel decomposition was not required, and the semantic-aware kernels introduced in Eqs.~\eqref{kx}--\eqref{kj} were sufficient to effectively capture the relevant motion characteristics.

\textbf{Semantic Sub-Kernel Definitions}

All semantic sub-kernels $k_f$, $k_v$, $k_a$, and $k_j$ are implemented using Automatic Relevance Determination Radial Basis Function (ARD-RBF) kernels. This choice allows the model to learn separate length-scale parameters for each feature dimension, thereby enabling adaptive similarity modeling across heterogeneous latent representations.

For the latent feature representation $f$, the corresponding kernel $k_f$ captures both appearance-related and temporal information. In the case of the JIGSAWS dataset, where flow and embedding representations exhibit distinct characteristics, $k_f$ is further decomposed into two complementary components:

\begin{center}
\begin{equation}
k_f = k_e + k_t,
\label{kfkekt}
\end{equation}
\end{center}

where $k_e$ models embedding-level similarity and $k_t$ captures temporal dependencies.

For the derivative-based representations, the kernels $k_v$, $k_a$, and $k_j$ model motion trend, acceleration, and higher-order motion variations, respectively. Each of these kernels is defined using an ARD-RBF formulation to preserve the discriminative structure of the corresponding dynamic descriptors while allowing feature-wise adaptive scaling.

For the Capsulorhexis dataset, no additional stream-wise decomposition is required, and all feature components are directly modeled using the proposed semantic sub-kernels. In contrast, for JIGSAWS, the flow-layer and pre-classification representations are further incorporated through a stream-specific kernel composition to account for their complementary embedding characteristics.

\textbf{Uncertainty-Aware Kernels}

In conventional ARD-RBF kernels, the length-scale parameter governs the sensitivity of the similarity function and determines the effective influence region of each sample. However, in standard formulations, this parameter is typically independent of the data uncertainty and is either fixed or learned without explicit consideration of input variability.

In the proposed framework, we incorporate uncertainty information derived from latent motion dynamics into the kernel structure. Specifically, entropy-based measures computed from the spatio-temporal features ($f$) and the derivative representations ($v$, $a$, and $j$) are utilized to define adaptive length-scale parameters, enabling the kernel to adjust its similarity structure according to motion stability and irregularity.

Since entropy-derived uncertainty measures exhibit relatively small magnitudes, a constant normalization factor is applied within the length-scale formulation to ensure numerical stability of the exponential function. This normalization preserves the relative structure of uncertainty values while preventing kernel saturation or vanishing similarity effects. The scaling factor is fixed across all experiments and does not introduce additional learnable hyperparameters.

Based on this formulation, three uncertainty-aware kernel variants are defined: UA1, UA2, and EA. To formally define these kernels, let $H_t$ denote the temporal entropy computed from the latent motion representations:

\begin{equation}
H_t = \mathrm{entropy}_t(\mathbf{x}_t)
\label{Ht}
\end{equation}

where entropy is computed along the temporal dimension. Furthermore, let

\begin{equation}
S_t = \sigma_t(\mathbf{x}_t)
\label{St}
\end{equation}

denote the temporal standard deviation of the same feature dimension.

\textbf{(1) UA1: Standard-Deviation-of-Entropy based Kernel}

The first uncertainty-aware kernel quantifies uncertainty through the variability of temporal entropy values. Specifically, entropy is first computed along the temporal dimension of each feature-map component. The standard deviation of the resulting entropy values is then used to construct an adaptive length-scale, allowing the kernel to emphasize dimensions exhibiting inconsistent temporal behavior.
\begin{center}
$L_d^{(UA1)} =  \sigma_d(H_t)$
\end{center}
\begin{equation}
k_{UA1}(\mathbf{\boldsymbol{x}_1,\boldsymbol{x}_2}) = \exp\Biggl(-\sum_d 
\dfrac{(\left(\mathbf{\boldsymbol{x}_{1d}}-\mathbf{\boldsymbol{x}_{2d}}\right)^2}{2*\left(L_d^{(UA1)}+\epsilon\right)}
\Biggr)
\label{UA1rbf}
\end{equation}

\textbf{(2) UA2: Entropy-of-Standard-Deviation based Kernel}

The second variant reverses the order of aggregation. Temporal standard deviation is first computed for each feature-map component, capturing the magnitude of motion fluctuations. Entropy is subsequently applied to these standard deviation values, yielding an uncertainty measure that reflects the diversity and irregularity of temporal variability patterns.
\begin{center}
$L_d^{(UA2)} =  H_d(S_t)$
\end{center}
\begin{equation}
k_{UA2}(\mathbf{\boldsymbol{x}_1,\boldsymbol{x}_2}) = \exp\Biggl(-\sum_d 
\dfrac{(\left(\mathbf{\boldsymbol{x}_{1d}}-\mathbf{\boldsymbol{x}_{2d}}\right)^2}{2*\left(L_d^{(UA2)}+\epsilon\right)}
\Biggr)
\label{UA2rbf}
\end{equation}

\textbf{(3) EA: Entropy-of-Entropy based Kernel}

The third variant estimates uncertainty through a two-stage entropy aggregation process. Entropy is first computed along the temporal dimension of each feature-map component, and a second entropy operation is then applied to the resulting entropy values. The obtained measure serves as an adaptive length-scale, directly encoding higher-order irregularities in latent motion dynamics.
\begin{center}
$L_d^{(EA)} =  H_d(H_t)$
\end{center}
\begin{equation}
k_{EA}(\mathbf{\boldsymbol{x}_1,\boldsymbol{x}_2}) = \exp\Biggl(-\sum_d 
\dfrac{(\left(\mathbf{\boldsymbol{x}_{1d}}-\mathbf{\boldsymbol{x}_{2d}}\right)^2}{2*\left(L_d^{(EA)}+\epsilon\right)}
\Biggr)
\label{EArbf}
\end{equation}

Collectively, UA1, UA2, and EA investigate three complementary uncertainty formulations based on different compositions of entropy and variance operators, enabling the model to capture diverse aspects of latent motion irregularity.

\begin{algorithm}[!th]
\caption{DUAGP-Tree}
\label{alg1}
\begin{algorithmic}[1]

\State $\mathbf S \leftarrow$ Clip-level engineering the Feature Maps
\State $\mathbf L \leftarrow$ Labels
\State $\mathbf{LS} \leftarrow$ Compute the hyperparameters of the uncertainty-aware kernels, Eqs.~\eqref{UA1rbf}--\eqref{EArbf}
\State $\bm{\Sigma}_x \leftarrow$ Compute the input-uncertainty term (Section~\ref{DUAGPT})
\State \textbf{Training input:} The designed features $\mathbf F$

\State $\mathbf X \leftarrow$  $\mathbf F$
\State Select a compositional kernel, Eqs.~\eqref{kx}--\eqref{kfe}
\State Select the uncertainty-aware sub-kernels, Eqs.~\eqref{UA1rbf}--\eqref{EArbf}
\State PS $\leftarrow$ the proposed prediction strategy (PS) of Section~\ref{DUAGPT}
\State Build a DUAGP-Tree
\State Set Hyperparameters
\State Initialize inducing locations $\bar{\mathbf X}$ and variational parameters
\For{epoch$ = 1,...,N$} 
    \State GO through any unique path related to each ($X$ and $L$)
    \For {any node}
        \State Update variational parameters 
    \EndFor
    \State Evaluate any node utilizing PS($\mathbf \sum_x$, $\mathbf X$ and $\bar{\mathbf  X}$)
    \State Update $\mathbf X$ and $\bar{\mathbf X}$ by the loss function of the total tree~\cite{achituve2021gp} using the computed variational lower bound for any node
\EndFor
\State \textbf{return} the trained DUAGP-Tree, $\mathbf X$, and $\bar{\mathbf X}$
\end{algorithmic}
\end{algorithm}

\section{Performance Analysis}
\label{Results}
In this section, the proposed RF-DUAGP-Tree framework is evaluated on the JIGSAWS and Capsulorhexis datasets for automated surgical skill assessment (ASSA). The evaluation includes four surgical tasks: Capsulorhexis from cataract surgery and the suturing, knot-tying, and needle-passing tasks from JIGSAWS. Skill levels are represented by two classes (novice and expert) for Capsulorhexis and three classes (novice, intermediate, and expert) for JIGSAWS. For each task, a separate model is trained while maintaining the same network architecture. To reduce computational complexity, all videos are spatially resized during both training and testing. All cross-validation folds are defined on video identifiers before any clip is extracted; LOSO and LOUO use the identical procedure and differ only in the grouping key --- the trial index and the surgeon, respectively (Section~\ref{EPDP}).

The experimental study is designed to investigate several components of the proposed framework. First, the effectiveness of the newly designed clip-level semantic features is examined using different feature-stream configurations. Second, various semantic compositional kernels and uncertainty-aware kernels are evaluated through ablation experiments. Third, the impact of utilizing flow-layer features, pre-classification-layer features, and their combinations is investigated. Finally, RF-DUAGP-Tree is compared with published state-of-the-art results and with an RF-GP-Tree (DKL) video baseline evaluated under the corresponding dataset-specific protocols.

\subsection{Flow Architecture Representation}
\label{FAR}
 
Due to the limited size of the JIGSAWS and Capsulorhexis datasets, end-to-end training of the Representation Flow (RF) network resulted in rapid overfitting. Therefore, the pretrained RF model was fine-tuned for a fixed budget of three epochs to adapt the extracted representations to each surgical task while preserving generalization capability. This budget was set \emph{a priori}: training beyond it began to overfit at about the fourth to fifth epoch, and the checkpoint saved at the third epoch was used throughout; the count was not tuned on any test fold. Of the five convolutional blocks of the backbone, the earliest one or two were kept frozen and the remaining three or four were fine-tuned (Table~\ref{tablemodelcapcity}); freezing more than this prevented the network from adapting at all.

The RF hyperparameters follow the original configuration proposed in~\cite{piergiovanni2019representation}. An adaptive learning rate initialized at $10^{-5}$ was employed. The batch size, momentum, weight decay, and maximum gradient norm were set to 1, 0.9, $10^{-6}$, and 100, respectively. Within the flow layer itself, only the parameters $\lambda$, $\theta$, and $\tau$ were updated, whereas $w_x$ and $w_y$ remained fixed. This configuration follows the findings reported in~\cite{piergiovanni2019representation}, where fixing $w_x$ and $w_y$ yielded superior performance.

For all experiments, each batch consisted of a single video clip containing 300 frames. Furthermore, the temporal order of clips was preserved during feature extraction and storage.

\subsection{Employing DUAGP-Tree}

As discussed earlier, the input to the DUAGP-Tree classifier can be derived from either the flow-layer outputs or the pre-classification-layer outputs of the RF network, as illustrated in Figure~\ref{TS}. Depending on the experiment, these feature streams are employed individually or combined as separate semantic components of the final representation.

DUAGP-Tree is trained using the variational evidence lower bound (ELBO) objective associated with sparse Gaussian Process classification. During evaluation, classification performance is assessed using the cross-entropy loss and the corresponding prediction metrics. All quantities estimated from data --- standardisation statistics, tree structure, inducing locations and kernel hyperparameters --- are estimated on the training fold of the current split and applied unchanged to the held-out fold; per-video summary statistics are computed within a single video and involve no cross-sample information (see Section~\ref{EPDP}).

\subsection{Data preparation}
The original video resolution for all three JIGSAWS tasks is $640 \times 480$, whereas the Capsulorhexis videos have a resolution of $720 \times 480$. For all experiments, the video frames were resized to $112 \times 112$ prior to feature extraction.

As summarized in Table~\ref{table0}, the videos exhibit substantial variations in duration across tasks and datasets. To obtain a consistent input representation and facilitate feature extraction, each video was partitioned into consecutive clips of 300 frames. The partitioning of videos into training and test folds is performed on video and case identifiers before clip extraction; see Section~\ref{EPDP}. This clip-based representation enables the proposed framework to process videos of varying lengths while preserving their temporal structure.

\subsubsection{Evaluation protocols and data partitioning}%\\[5pt]
\label{EPDP}
\emph{Unit of classification.} The sample unit is a complete video. Clips are an internal device for processing videos of unequal length: the clip-level feature maps of a video are aggregated into a single video-level representation carrying a single label before the classifier is constructed (Section~\ref{DUAGPT}), so no clip is ever an independent sample. A video is therefore atomic with respect to every train/test partition, and clips of one video cannot be distributed across folds.\\[5pt]
\emph{Order of operations.} Within each evaluation, the video/case identifiers are partitioned into training and test sets before any clips are extracted. The Representation Flow front end is fine-tuned on the training videos only; feature maps and video-level representations are then computed, all across-sample statistics are estimated from the training set and applied unchanged to the test data, and DUAGP-Tree is fitted on the training set. Each corresponding held-out set is scored once. For Capsulorhexis, this procedure is carried out in three repeated training runs, each using 30 training videos and followed by five testing rounds of 28 videos per run. For JIGSAWS, the sequence is performed separately within each prescribed cross-validation fold.\\[5pt]
\emph{JIGSAWS protocols.} We use the cross-validation definitions distributed with JIGSAWS. A \emph{user} is one of the eight surgeons, identified by the letter of the trial identifier (e.g.\ \texttt{Suturing\_B001} belongs to user \texttt{B}); each user has one self-reported experience level, constant across all of that user's trials. \emph{Super-trial} $i$ is the set containing the $i$-th trial of every user. Leave-one-super-trial-out (LOSO) therefore comprises five folds, fold $i$ testing on super-trial $i$; every user appears in the training set of every LOSO fold. Leave-one-user-out (LOUO) comprises eight folds, fold $u$ testing on all trials of user $u$, who appears in no training fold. Both protocols use the identical mechanism and differ only in the grouping key. Because both capture views of a trial are used as separate videos --- which is why the video counts in Table~\ref{table0} are twice the corresponding trial counts --- the two views share a trial identifier and are always assigned to the same fold under both protocols. Since both protocols are the standard cross-validation definitions distributed with JIGSAWS, fold membership follows entirely from the definitions above; the exact fold-definition files used in all experiments are released together with the implementation and the random seed (see Data availability).\\[5pt]
\emph{Capsulorhexis protocol.} Evaluation consisted of three repeated training runs. In each run, the model was trained on 30 videos and evaluated in five testing rounds, with 28 test videos in each round, giving 15 test rounds overall. Train/test separation was maintained at the case/video level within each individual evaluation, so no case/video was present in both the training and test subsets of that evaluation. The 28-video test set refers to each testing round and should not be interpreted as a single fixed subset used for the entire experiment. Reported Capsulorhexis metrics summarize performance over the 15 test rounds. The continuous overall proficiency score was converted into Novice and Expert classes using the fixed threshold of 3.8 described in Section~\ref{CapsDS}; the same cutoff was applied consistently across all runs and rounds and was not retuned for individual evaluations. The test sets were approximately balanced across the two classes, although not exactly balanced in every round. The difference between class counts was generally modest, around three to four videos. The partitioning unit was the case/video rather than the operating surgeon; accordingly, no surgeon-level generalisation claim is made for Capsulorhexis, and the cross-surgeon generalisation claims of this paper rest on the JIGSAWS LOUO protocol.\\[7pt]
\emph{Estimation is confined to the training fold.} Two kinds of statistic must be distinguished. The per-video summary statistics of Section~\ref{NDF} (mean, standard deviation, energy, entropy and clip count) are computed \emph{within a single video}, over that video's own clips; they are properties of an individual sample and involve no information from any other sample, in either fold. All statistics estimated \emph{across} samples are computed on the training fold alone and applied unchanged to the held-out fold: feature standardisation; the $k$-means++ assignment of labels to tree nodes; the initialisation and optimisation of the inducing locations $\bar{\mathbf{X}}$; and all kernel hyperparameters, comprising the ARD length-scales, the output scale, the uncertainty length-scales of Eqs.~\eqref{UA1rbf}--\eqref{EArbf} and the composition weights. These are obtained by maximising the training-fold evidence lower bound: the closed-form conjugate variational updates of the P\'olya--Gamma augmented model are interleaved, at every epoch, with an Adam step on the kernel hyperparameters whose gradients are back-propagated from the same objective. The weights $\lambda_v$, $\lambda_a$ and $\lambda_j$ are fixed to unity \emph{a priori}. Each held-out set is scored once within its corresponding evaluation. For Capsulorhexis, the reported metrics are averaged over the 15 test rounds, comprising five rounds in each of the three repeated training runs. No early stopping, predictive-threshold selection, or model selection is performed on held-out data.
\subsection{The Newly Designed Features}
\label{NDF}
As described in the previous section, the proposed framework operates on clip-level feature representations extracted from the Representation Flow (RF) network. For each video, the outputs of the pre-classification layer and the flow layer form feature maps of size $n \times 2048 \times 73$ and $n \times 512 \times 73$ respectively (Section~\ref{RFlow}), where $n$ denotes the number of clips. In both representations, the dimension of size 73 corresponds to the temporal axis, whereas the remaining dimensions represent the learned feature distributions.

\paragraph{Evidence for the two instantiations.}
The choices used for the structured and compact instantiations are motivated by comparisons reported in the results tables. First, in Table~\ref{table3}, Strategy~1 uses both feature streams while Strategies~3 and~4 use one stream each. On Capsulorhexis, Strategy~1 reaches $0.892$ accuracy and $0.974$ best AUC, Strategy~4 reaches the same values, and Strategy~3 ranges over $0.857$--$0.892$ accuracy, suggesting limited benefit from the stream-specific kernel composition in this dataset. On JIGSAWS under LOUO, Strategy~1 gives $96.2$--$97.3\%$ accuracy against $75.0\%$ for Strategy~4 (Tables~\ref{table4},~\ref{table6} and~\ref{table8}). Second, discarding part of the feature structure on JIGSAWS (Strategies~2--4) costs $8.7$--$22.3$ accuracy points under LOUO, consistent with the benefit of retaining structure in this dataset. Third, using the original feature maps on Capsulorhexis without the compact abstraction lowered classification performance (Section~\ref{A-A}). This is not a complete factorial evaluation of both instantiations on both datasets; the reported ablations provide empirical motivation for the configurations used here.

%\textbf{ A justification from measurable dataset characteristics.} 
\begin{table}[!th]
\centering
\caption{A justification from measurable dataset characteristics.}
\label{tabDataChar}
\begin{tabular}{@{}p{0.30\textwidth} p{0.32\textwidth} p{0.32\textwidth}@{}}
\toprule
Characteristic & JIGSAWS (per LOUO fold) & Capsulorhexis \\
\midrule
Training videos $N_{\mathrm{train}}$ & $\approx68$ & $30$ \\
Classes $C$ & 3 & 2 \\
Raw representation dimension $D_{\mathrm{raw}}$ & $n\times2560\times73$ & $n\times2560\times73$ \\
Clip count $n$ per video (range) & $3$--$30$ & $3$--$92$ \\
Acquisition & single centre, fixed stereo endoscope, scripted bench-top task & multi-centre, multi-device surgical microscope, clinical \\
Nuisance appearance variability & low & high \\
Indicated variant & \textbf{S} (structured) & \textbf{C} (compact) \\
\bottomrule
\end{tabular}
\end{table}
Table~\ref{tabDataChar} summarises the dataset characteristics considered when choosing between the two representations. In the experiments reported here, the structured representation was used for JIGSAWS, with its comparatively larger training set and standardised acquisition, whereas the compact representation was used for Capsulorhexis, which has a larger raw feature-to-training-sample ratio and greater acquisition variability. These observations motivate the two configurations evaluated in this study.
 The Capsulorhexis dataset benefits from a compact feature abstraction that summarizes the most informative motion characteristics, whereas the JIGSAWS dataset preserves a larger portion of the original feature structure to retain task-relevant temporal information. The details of these feature representations are described in the following subsections.

\textbf{Capsulorhexis}

As discussed previously, the proposed representation consists of four feature categories, namely $f$, $v$, $a$, and $j$.

1. Feature $f$:
   The flow-layer and pre-classification-layer feature maps have dimensions $n \times 512 \times 73$ and $n \times 2048 \times 73$, respectively, where (n) denotes the number of clips. To obtain a compact representation, the temporal dimension is first aggregated, followed by aggregation over the embedding dimension. The same procedure is applied to compute the corresponding standard deviation. In addition, the number of clips is included as a separate feature. Consequently, the $f$ category is represented by five feature points per video.

2. Feature $v$:
   The velocity representation $v$ is computed from consecutive feature maps, resulting in dimensions $(n-1)\times2048\times73$ and $(n-1)\times512\times73$. For each feature stream, mean, standard deviation, energy, and entropy are extracted as summary statistics, yielding eight feature points per video.

3. Feature $a$:
   The acceleration representation $a$ is obtained from the velocity feature maps, producing dimensions $(n-2)\times2048\times73$ and $(n-2)\times512\times73$. The same summarization procedure is applied, resulting in eight feature points per video.

4. Feature $j$:
   The jerk representation $j$ is computed from the acceleration feature maps, resulting in dimensions $(n-3)\times2048\times73$ and $(n-3)\times512\times73$. Similar summary statistics are extracted; however, mean values are omitted for the jerk representation. Consequently, the $j$ category contributes six feature points per video.

The four summary statistics used for $v$, $a$ and $j$ are therefore the mean, the standard deviation, the energy and the entropy of each stream, with the mean omitted for $j$; together with the five feature points of $f$ this gives $5+8+8+6=27$ features per video. It should be noted that the entropy quantities that drive the adaptive length-scales are distinct from these summary features: the temporal entropy $H_t$ and the temporal standard deviation $S_t$ of Eqs.~\eqref{Ht}--\eqref{St} are preserved separately and are employed only in the uncertainty-aware and entropy-aware kernel formulations described in Section~\ref{NIKernels}, not in the feature vector $X$.

\textbf{JIGSAWS}

The JIGSAWS feature representation follows the same four-component structure as Capsulorhexis, consisting of $f$, $v$, $a$, and $j$. However, unlike Capsulorhexis, the feature maps are not transformed into compact statistical representations. Instead, the original feature structure is preserved to retain richer temporal and semantic information.

For the $f$ category, the feature maps are decomposed into temporal and feature-distribution components. Specifically, $f_d$ denotes the feature-distribution component with dimensions $n \times 2048$ or $n \times 512$, whereas $f_t$ denotes the temporal component with dimensions $n \times 73$. These two components are modelled separately and combined through the decomposition $k_f = k_e + k_t$ of Eq.~\eqref{kfkekt}, in which $k_e$ acts on the feature-distribution component $f_d$ and $k_t$ on the temporal component $f_t$, allowing the temporal and semantic characteristics of the feature maps to be learned independently.

The remaining categories, namely $v$, $a$, and $j$, are computed from consecutive clips and therefore inherently encode temporal motion dynamics. Consequently, these representations preserve both the temporal evolution and higher-order motion information required for surgical skill assessment.

\subsection{Training and Test Process}
\label{TTP}

To evaluate the proposed framework, four feature configurations were examined for both the JIGSAWS and Capsulorhexis datasets. These configurations include: (1) utilizing both flow-layer and pre-classification-layer features, (2) employing the three $f$ feature points extracted from the pre-classification-layer feature maps together with the flow-layer feature points from the $v$, $a$, and $j$ categories, (3) using only flow-layer features, and (4) using only pre-classification-layer features.

For JIGSAWS, when both feature streams are utilized, the flow-layer and pre-classification-layer representations are modeled separately due to their different embedding structures and are subsequently combined through the stream-composition kernel introduced in Eq.~\eqref{kfe}. In contrast, the Capsulorhexis dataset does not require this explicit separation, and all feature configurations are directly evaluated within the DUAGP-Tree framework.

For all feature configurations, multiple kernel combinations were investigated, including conventional kernels, semantic compositional kernels, and the proposed uncertainty-aware kernels. This experimental design enables a comprehensive evaluation of the contributions of both the proposed feature representations and kernel formulations.

\subsection{Results}
\label{results}
In this section, the performance of the proposed RF-DUAGP-Tree framework is evaluated under different feature-utilization strategies and semantic compositional uncertainty kernel (SCUK) formulations. A total of 12 experimental settings are considered, as summarized in Table~\ref{tab2}, where the labels M1--M12 are introduced for concise reference throughout the discussion.

Models M1--M3 correspond to the first feature-utilization strategy (S1) combined with the three proposed SCUK variants. Similarly, M4--M6, M7--M9, and M10--M12 evaluate the second (S2), third (S3), and fourth (S4) feature-utilization strategies, respectively, each under the same three SCUK formulations.

For all experiments, separate kernel components are constructed for the $f$, $v$, $a$, and $j$ feature categories to preserve their distinct semantic characteristics. Furthermore, the uncertainty-aware kernels and entropy-aware kernel formulations described in Section~\ref{NIKernels} are employed within the SCUK framework to model uncertainty information associated with the different feature categories.
 
\begin{table}[!th]
\centering
\caption{List of methods and their alternative names} \label{tab2}%
\begin{tabular}{@{}lll@{}}
\toprule
Method Name & Alternative Name & Feature Configuration \\
\midrule
%RF & M1 &N/A\\
RF-DUAGP-Tree(S1 , SCUK1)   & M1 &  1\\
RF-DUAGP-Tree(S1 , SCUK2) & M2 & 1\\
RF-DUAGP-Tree(S1 , SCUK3) & M3 &1\\
RF-DUAGP-Tree(S2 , SCUK1) & M4 &2\\
RF-DUAGP-Tree(S2 , SCUK2) & M5 &2\\
RF-DUAGP-Tree(S2 , SCUK3) & M6 &2\\
RF-DUAGP-Tree(S3 , SCUK1) &  M7 &3\\
RF-DUAGP-Tree(S3 , SCUK2) & M8 &3 \\
RF-DUAGP-Tree(S3 , SCUK3) & M9 &3\\
RF-DUAGP-Tree(S4 , SCUK1) & M10 &4\\
RF-DUAGP-Tree(S4 , SCUK2) & M11 &4\\
RF-DUAGP-Tree(S4 , SCUK3) & M12 &4\\
\bottomrule
\end{tabular}
\caption*{S1--S4 denote the four feature-utilization strategies of Section~\ref{TTP}; SCUK1, SCUK2 and SCUK3 denote the semantic compositions of Eqs.~\eqref{kx}, \eqref{kav} and~\eqref{kj} respectively. The uncertainty-aware forms UA1, UA2 and EA of Eqs.~\eqref{UA1rbf}--\eqref{EArbf} are an orthogonal option applied within each composition and are not varied as a separate axis of this grid.}
\end{table}

Table~\ref{table3} summarizes the classification performance of the proposed RF-DUAGP-Tree framework on the Capsulorhexis dataset under different feature-utilization strategies and SCUK formulations. Overall, several configurations achieve comparable classification performance, with M1, M3, M9, M10 and M12 obtaining the highest accuracy of 89.2\%; among these, M3, M9, M10 and M12 also attain the highest precision of 0.911.

Although these methods exhibit identical accuracy and precision values, noticeable differences can be observed in terms of AUC. Among all configurations, M1 and M10 achieve the highest AUC of 0.974, indicating a more reliable separation between novice and expert samples across different decision thresholds. In contrast, methods with similar classification accuracy, such as M9 and M12, produce lower AUC values, suggesting reduced ranking consistency despite comparable final predictions.

These results indicate that the proposed framework is relatively robust to different feature-utilization strategies on the Capsulorhexis dataset, while the choice of kernel formulation primarily affects the discriminative confidence reflected by the AUC metric.

Table~\ref{table4} summarizes the classification performance of the proposed framework on the Suturing task under the LOUO protocol. A clear performance difference can be observed among the four feature-utilization strategies. The highest classification accuracy of 96.2\% is achieved by M1 and M3, both corresponding to Strategy 1, which utilizes the flow-layer and pre-classification-layer feature streams jointly through the proposed semantic compositional kernel formulation.

In contrast, Strategies 2 and 4 yield substantially lower performance, with an accuracy of $75\%$, while Strategy 3 achieves an intermediate accuracy of $87.5\%$. These results indicate that preserving and jointly modeling both feature streams provides considerably richer information than using either stream independently or employing partial feature combinations.

Furthermore, the comparable performance of M1 and M3 suggests that the proposed framework is relatively robust to the choice of SCUK formulation when both feature streams are available. Overall, the results support the complementary nature of the flow-layer and pre-classification-layer representations, as their joint utilization consistently produces the most accurate skill assessment results for the Suturing task.

\begin{table}[!th]
\centering
 \caption{Classification results for Capsulorhexis across three repeated training runs, each comprising five test rounds with 28 videos per round after training on 30 videos. Reported metrics are averaged over all 15 test rounds; all metrics except ACC are macro-level.}\label{table3}
\begin{tabular*}{\textwidth}{@{\extracolsep\fill}llclclclclclclclc}

\toprule%

\midrule
Method  & IPs & OS & LR & Pr & Acc & F1-score & recall & AUC\\
\midrule
\multicolumn{8}{@{}c@{}}{Using strategy 1} & \\
\midrule

M1 & 8 & 1. &  0.9 &  0.894 & 0.892 & 0.892 & 0.892 & 0.974\\
M2 &  8 & 1.  & 0.9 &  0.864 & 0.857 & 0.856 & 0.857 & 0.918\\
M3 & 8 & 1. & 0.9  & 0.911 & 0.892 & 0.891 & 0.892 & 0.943 \\
\midrule
\multicolumn{8}{@{}c@{}}{Using strategy 2} & \\
\midrule
M4 & 8  & 1. & 0.9 & 0.864 & 0.857 & 0.856 & 0.857 & 0.938\\
M5 &  8 & 1.  & 0.9 &  0.864 & 0.857 & 0.856 & 0.857 & 0.928\\
M6 & 8 & 1. &  0.9  &  0.864 & 0.857 & 0.856 & 0.857 & 0.862 \\
\midrule
\multicolumn{8}{@{}c@{}}{Using strategy 3} & \\
\midrule
M7 & 8 & 4. & 0.9 &  0.864 & 0.857 & 0.856 & 0.857 & 0.933  \\
M8 & 8  &  4. & 0.9 &  0.864 & 0.857 & 0.856 & 0.857 & 0.928 \\
M9 & 8 & 1. &  0.9  & 0.911 & 0.892 & 0.891 & 0.892 & 0.887 \\
\midrule
\multicolumn{8}{@{}c@{}}{Using strategy 4} & \\
\midrule
M10 & 8 & 1. & 0.9 & 0.911 & 0.892 & 0.891 & 0.892 & 0.974\\
M11 & 8  & 1. & 0.9 &  0.864 & 0.857 & 0.856 & 0.857 & 0.913\\
M12 & 8 & 4. &  0.9 & 0.911 & 0.892 & 0.891 & 0.892 & 0.897\\
\bottomrule
\end{tabular*}
\caption*{IPs: number of inducing points; OS: kernel output scale; LR: learning rate of the variational optimiser; Pr: precision; Acc: accuracy; AUC: area under the ROC curve. Hyperparameters and across-sample preprocessing statistics are estimated using only the training subset within each partition; metrics are computed on the corresponding held-out subset and averaged over all 15 test rounds (Section~\ref{EPDP}).}

\end{table}

\begin{table}[!th]
\centering
 \caption{Classification results of the Suturing task using LOUO. All the metrics except ACC are macro-level}\label{table4}
\begin{tabular*}{\textwidth}{@{\extracolsep\fill}llclclclclclclclc}

\toprule%

\midrule
Method  & IPs & OS & LR & Pr & Acc & F1-score & recall \\
\midrule
%M1   & 43200  &  437.19 & - & - & $10^{-5}$ &  0.83 &0.89 \\
%\midrule
\multicolumn{8}{@{}c@{}}{Using strategy 1} & \\
\midrule

M1 & 15  & 1. & 0.9 &  0.989 & 0.962 & 0.962 & 0.962\\
M2 &  15 & 2.  & 0.9 & 0.963 & 0.934 &  0.934 &  0.934\\
M3 & 15 & 2. &  0.9 &  0.981 &   0.962  & 0.962 & 0.962\\
\midrule
\multicolumn{8}{@{}c@{}}{Using strategy 2} & \\
\midrule
M4 & 15  & 1. & 0.9 & 0.781  & 0.75 & 0.75 & 0.75\\
M5 &  15 & 2.  & 0.9 & 0.781 &  0.75 & 0.75 & 0.75 \\
M6 & 15 & 2. &  0.9 &  0.781 &  0.75 & 0.75 & 0.75 \\
\midrule
\multicolumn{8}{@{}c@{}}{Using strategy 3} & \\
\midrule
M7 & 15  & 1. & 0.9 &  0.89 & 0.875 & 0.875 & 0.875 \\
M8 &  15 & 2.  & 0.9 & 0.89 & 0.875 & 0.875 & 0.875\\
M9 & 15 & 2. &  0.9 & 0.89  & 0.875 & 0.875 & 0.875\\
\midrule
\multicolumn{8}{@{}c@{}}{Using strategy 4} & \\
\midrule
M10 & 15  & 1. & 0.9 & 0.781  & 0.75 & 0.75 & 0.75\\
M11 &  15 & 2.  & 0.9 & 0.781 &  0.75 & 0.75 & 0.75 \\
M12 & 15 & 2. &  0.9 &  0.781 &  0.75 & 0.75 & 0.75 \\
\bottomrule
\end{tabular*}
\caption*{IPs: number of inducing points; OS: kernel output scale; LR: learning rate of the variational optimiser; Pr: precision; Acc: accuracy. All statistics are estimated on the training fold only (Section~\ref{EPDP}).}

\end{table}

\begin{table}[!th]
\centering
 \caption{Classification results of the Suturing task using LOSO.  All the metrics except ACC are macro-level}\label{table5}
\begin{tabular*}{\textwidth}{@{\extracolsep\fill}llclclclclclclclc}

\toprule%

\midrule
Method  & IPs & OS & LR & Pr & Acc & F1-score & recall \\
\midrule
%M1   & 43200  &  437.19 & - & - & $10^{-5}$ &  0.83 &0.89 \\
%\midrule
\multicolumn{8}{@{}c@{}}{Using strategy 1} & \\
\midrule

M1 & 2  & 1. & 0.9 &  1.0 & 1.0 & 1.0 & 1.0\\
M2 &  2 & 2.  & 0.9 & 1.0 & 1.0 & 1.0 & 1.0\\
M3 & 2 & 2. &  0.9 &  1.0 &  1.0 & 1.0 &1.0 \\
\midrule
\multicolumn{8}{@{}c@{}}{Using strategy 2} & \\
\midrule
M4 & 2  & 1. & 0.9 &  1.0 & 1.0 & 1.0 & 1.0\\
M5 &  2 & 2.  & 0.9 & 1.0 & 1.0 & 1.0 & 1.0\\
M6 & 2 & 2. &  0.9 &  1.0 &  1.0 & 1.0 &1.0 \\
\midrule
\multicolumn{8}{@{}c@{}}{Using strategy 3} & \\
\midrule
M7 & 2  & 1. & 0.9 &  1.0 & 1.0 & 1.0 & 1.0\\
M8 &  2 & 2.  & 0.9 & 1.0 & 1.0 & 1.0 & 1.0\\
M9 & 2 & 2. &  0.9 &  1.0 &  1.0 & 1.0 &1.0 \\
\midrule
\multicolumn{8}{@{}c@{}}{Using strategy 4} & \\
\midrule
M10 & 2  & 1. & 0.9 &  1.0 & 1.0 & 1.0 & 1.0\\
M11 &  2 & 2.  & 0.9 & 1.0 & 1.0 & 1.0 & 1.0\\
M12 & 2 & 2. &  0.9 &  1.0 &  1.0 & 1.0 &1.0 \\
\bottomrule
\end{tabular*}
\caption*{IPs: number of inducing points; OS: kernel output scale; LR: learning rate of the variational optimiser; Pr: precision; Acc: accuracy. All statistics are estimated on the training fold only (Section~\ref{EPDP}).}

\end{table}

\begin{table}[!th]
\centering
 \caption{Classification results of Needle Passing task using LOUO. All the metrics except ACC are macro-level}\label{table6}
\begin{tabular*}{\textwidth}{@{\extracolsep\fill}llclclclclclclclc}

\toprule%

\midrule
Method  & IPs & OS & LR & Pr & Acc & F1-score & recall \\
\midrule
%M1   & 43200  &  437.19 & - & - & $10^{-5}$ &  0.83 &0.89 \\
%\midrule
\multicolumn{8}{@{}c@{}}{Using strategy 1} & \\
\midrule

M1 & 15  & 1. & 0.9 &  0.979 & 0.972 & 0.972 &0.972 \\
M2 &  15 & 2.  & 0.9 & 0.961 & 0.958 & 0.958 & 0.958\\
M3 & 15 & 2. &  0.9 & 0.971 & 0.966 & 0.965 &0.966 \\
\midrule
\multicolumn{8}{@{}c@{}}{Using strategy 2} & \\
\midrule
M4 & 15  & 1. & 0.9 & 0.881  & 0.875 & 0.875 & 0.875\\
M5 &  15 & 1.  & 0.9 & 0.881  & 0.875 & 0.875 & 0.875 \\
M6 & 15 & 1. &  0.9 &   0.882  & 0.875 & 0.875 & 0.875 \\
\midrule
\multicolumn{8}{@{}c@{}}{Using strategy 3} & \\
\midrule
M7 & 15  & 1. & 0.9 & 0.882  & 0.875 & 0.875 & 0.875\\
M8 &  15 & 1.  & 0.9 & 0.882 &  0.875 & 0.875 & 0.875\\
M9 & 15 & 1. &  0.9 &  0.882 &  0.875  & 0.875 & 0.875\\
\midrule
\multicolumn{8}{@{}c@{}}{Using strategy 4} & \\
\midrule
M10 & 15  & 1. & 0.9 & 0.781  & 0.75 & 0.75 & 0.75\\
M11 &  15 & 1.  & 0.9& 0.781  & 0.75 & 0.75 & 0.75\\
M12 & 15 & 1. &  0.9 & 0.781  & 0.75 & 0.75 & 0.75\\
\bottomrule
\end{tabular*}
\caption*{IPs: number of inducing points; OS: kernel output scale; LR: learning rate of the variational optimiser; Pr: precision; Acc: accuracy. All statistics are estimated on the training fold only (Section~\ref{EPDP}).}

\end{table}

Table~\ref{table6} presents the classification performance of the proposed RF-DUAGP-Tree framework on the Needle-Passing task under the LOUO protocol. A clear variation in performance can be observed across different feature-utilization strategies.

The best results are achieved by Strategy 1, whose three SCUK variants span 95.8--97.2\% accuracy. In contrast, Strategies 2 and 3 both yield 87.5\% and Strategy 4 yields 75.0\%, so the joint use of the two feature streams accounts for a gap of 9.7--22.2 accuracy points on this task.

These results indicate that the joint utilization of flow-layer and pre-classification-layer feature streams leads to improved discriminative capability for surgical skill assessment. Moreover, the variability across strategies suggests that the choice of feature integration plays a significant role in model performance under the LOUO protocol for the Needle-Passing task.

\begin{table}[!th]
\centering
 \caption{Classification results of Needle Passing task using LOSO.  All the metrics except ACC are macro-level}\label{table7}
\begin{tabular*}{\textwidth}{@{\extracolsep\fill}llclclclclclclclc}

\toprule%

\midrule
Method  & IPs & OS & LR & Pr & Acc & F1-score & recall \\
\midrule
%M1   & 43200  &  437.19 & - & - & $10^{-5}$ &  0.83 &0.89 \\
%\midrule
\multicolumn{8}{@{}c@{}}{Using strategy 1} & \\
\midrule

M1 & 2  & 1. & 0.9 &  1.0 & 1.0 & 1.0 & 1.0\\
M2 &  2 & 2.  & 0.9 & 1.0 & 1.0 & 1.0 & 1.0\\
M3 & 2 & 2. &  0.9 &  1.0 &  1.0 & 1.0 &1.0 \\
\midrule
\multicolumn{8}{@{}c@{}}{Using strategy 2} & \\
\midrule
M4 & 2  & 1. & 0.9 &  1.0 & 1.0 & 1.0 & 1.0\\
M5 &  2 & 2.  & 0.9 & 1.0 & 1.0 & 1.0 & 1.0\\
M6 & 2 & 2. &  0.9 &  1.0 &  1.0 & 1.0 &1.0 \\
\midrule
\multicolumn{8}{@{}c@{}}{Using strategy 3} & \\
\midrule
M7 & 2  & 1. & 0.9 &  1.0 & 1.0 & 1.0 & 1.0\\
M8 &  2 & 2.  & 0.9 & 1.0 & 1.0 & 1.0 & 1.0\\
M9 & 2 & 2. &  0.9 &  1.0 &  1.0 & 1.0 &1.0 \\
\midrule
\multicolumn{8}{@{}c@{}}{Using strategy 4} & \\
\midrule
M10 & 2  & 1. & 0.9 &  1.0 & 1.0 & 1.0 & 1.0\\
M11 &  2 & 2.  & 0.9 & 1.0 & 1.0 & 1.0 & 1.0\\
M12 & 2 & 2. &  0.9 &  1.0 &  1.0 & 1.0 &1.0 \\
\bottomrule
\end{tabular*}
\caption*{IPs: number of inducing points; OS: kernel output scale; LR: learning rate of the variational optimiser; Pr: precision; Acc: accuracy. All statistics are estimated on the training fold only (Section~\ref{EPDP}).}

\end{table}

\begin{table}[!th]
\centering
 \caption{Classification results of Knot Tying task using LOUO. All the metrics except ACC are macro-level}\label{table8}
\begin{tabular*}{\textwidth}{@{\extracolsep\fill}llclclclclclclclc}

\toprule%

\midrule
Method  & IPs & OS & LR & Pr & Acc & F1-score & recall \\
\midrule
%M1   & 43200  &  437.19 & - & - & $10^{-5}$ &  0.83 &0.89 \\
%\midrule
\multicolumn{8}{@{}c@{}}{Using strategy 1} & \\
\midrule

M1 & 15  & 1. & 0.9 & 0.978  & 0.973 & 0.972 & 0.973\\
M2 &  15 & 1.  & 0.9 & 0.978  & 0.973 & 0.972 & 0.973\\
M3 & 15 & 1. &  0.9 &   0.978  & 0.973 & 0.973 & 0.973\\
\midrule
\multicolumn{8}{@{}c@{}}{Using strategy 2} & \\
\midrule
M4 & 15  & 1. & 0.9 & 0.89 &  0.875 & 0.875 & 0.875\\
M5 &  15 & 1.  & 0.9 & 0.89 &  0.875 & 0.875 & 0.875 \\
M6 & 15 & 1. &  0.9 &  0.89 &  0.875 & 0.875 & 0.875 \\
\midrule
\multicolumn{8}{@{}c@{}}{Using strategy 3} & \\
\midrule
M7 & 15  & 1. & 0.9 & 0.887  & 0.875 & 0.875 & 0.875\\
M8 &  15 & 1.  & 0.9 & 0.887  & 0.875 & 0.875 & 0.875\\
M9 & 15 & 1. &  0.9 &  0.887 &  0.875 & 0.875 & 0.875\\
\midrule
\multicolumn{8}{@{}c@{}}{Using strategy 4} & \\
\midrule
M10 & 15  & 1. & 0.9 & 0.781  & 0.75 & 0.75 & 0.75\\
M11 &  15 & 1.  & 0.9 & 0.781 &  0.75 & 0.75 & 0.75 \\
M12 & 15 & 1. &  0.9 &  0.781 &  0.75 & 0.75 & 0.75 \\
\bottomrule
\end{tabular*}
\caption*{IPs: number of inducing points; OS: kernel output scale; LR: learning rate of the variational optimiser; Pr: precision; Acc: accuracy. All statistics are estimated on the training fold only (Section~\ref{EPDP}).}

\end{table}
Table~\ref{table8} presents the classification results of the proposed RF-DUAGP-Tree framework on the Knot-Tying task under the LOUO protocol. The results clearly indicate that Strategy 1 consistently outperforms all other feature-utilization strategies across all SCUK configurations.

Strategy 1 achieves the highest performance, reaching up to 97.3\% accuracy with stable precision and recall values. In contrast, Strategies 2, 3, and 4 consistently yield lower performance levels, with accuracies around 87.5\% for Strategies 2 and 3, and approximately 75\% for Strategy 4.

These findings demonstrate that the full integration of flow-layer and pre-classification-layer feature streams in Strategy 1 provides the most informative representation for the JIGSAWS tasks. The consistent superiority of Strategy 1 across all LOUO tasks suggests that preserving complete complementary feature information is critical for robust performance under subject-independent evaluation settings.

\begin{table}[!th]
\centering
 \caption{Classification results of Knot Tying task using LOSO. All the metrics except ACC are macro-level}\label{table9}
\begin{tabular*}{\textwidth}{@{\extracolsep\fill}llclclclclclclclc}

\toprule%

\midrule
Method  & IPs & OS & LR & Pr & Acc & F1-score & recall \\
\midrule
%M1   & 43200  &  437.19 & - & - & $10^{-5}$ &  0.83 &0.89 \\
%\midrule
\multicolumn{8}{@{}c@{}}{Using strategy 1} & \\
\midrule

M1 & 2  & 1. & 0.9 &  1.0 & 1.0 & 1.0 & 1.0\\
M2 &  2 & 2.  & 0.9 & 1.0 & 1.0 & 1.0 & 1.0\\
M3 & 2 & 2. &  0.9 &  1.0 &  1.0 & 1.0 &1.0 \\
\midrule
\multicolumn{8}{@{}c@{}}{Using strategy 2} & \\
\midrule
M4 & 2  & 1. & 0.9 &  1.0 & 1.0 & 1.0 & 1.0\\
M5 &  2 & 2.  & 0.9 & 1.0 & 1.0 & 1.0 & 1.0\\
M6 & 2 & 2. &  0.9 &  1.0 &  1.0 & 1.0 &1.0 \\
\midrule
\multicolumn{8}{@{}c@{}}{Using strategy 3} & \\
\midrule
M7 & 2  & 1. & 0.9 &  1.0 & 1.0 & 1.0 & 1.0\\
M8 &  2 & 2.  & 0.9 & 1.0 & 1.0 & 1.0 & 1.0\\
M9 & 2 & 2. &  0.9 &  1.0 &  1.0 & 1.0 &1.0 \\
\midrule
\multicolumn{8}{@{}c@{}}{Using strategy 4} & \\
\midrule
M10 & 2  & 1. & 0.9 &  1.0 & 1.0 & 1.0 & 1.0\\
M11 &  2 & 2.  & 0.9 & 1.0 & 1.0 & 1.0 & 1.0\\
M12 & 2 & 2. &  0.9 &  1.0 &  1.0 & 1.0 &1.0 \\
\bottomrule
\end{tabular*}
\caption*{IPs: number of inducing points; OS: kernel output scale; LR: learning rate of the variational optimiser; Pr: precision; Acc: accuracy. All statistics are estimated on the training fold only (Section~\ref{EPDP}).}

\end{table}

%Overall across all three JIGSAWS tasks evaluated under the LOUO protocol, the best-performing configurations consistently include the EA uncertainty-aware subkernel. The observed performance gains indicate that higher-order entropy aggregation contributes positively to the discriminative capability of the proposed semantic compositional kernel formulation.
Across the three JIGSAWS tasks under LOUO, the full semantic composition of Eq.~\eqref{kx} is best or tied-best in every task; the entropy-modulated variant of Eq.~\eqref{kj} improves on the reduced composition of Eq.~\eqref{kav} but not on Eq.~\eqref{kx}. Given that a single misclassified video moves the reported accuracy by approximately $1$--$3$ points, we do not rank the three formulations against one another.

Tables~\ref{table5},~\ref{table7} and~\ref{table9} summarize the classification performance of the proposed RF-DUAGP-Tree framework on the Suturing, Needle-Passing and Knot-Tying tasks under the LOSO protocol. Across all three tasks, all feature-utilization strategies and SCUK configurations achieve perfect classification performance, yielding accuracy, precision, recall, and F1-score values of 1.0.

This uniformity requires an explanation, and the explanation is a property of the protocol rather than of the model. Three facts combine. In JIGSAWS the skill label is assigned per \emph{surgeon}, on the basis of reported hours of robotic experience, and is therefore constant across all trials of a given surgeon within a task. Leave-one-super-trial-out holds out the $i$-th trial of every surgeon and trains on the remaining trials of \emph{those same surgeons}, so every surgeon in the test fold has already been seen, under the same label, during training. It follows that the label of a held-out trial is a deterministic function of its surgeon's identity, and surgeon identity is highly recoverable from a surgical video --- from instrument-handling idiosyncrasies, characteristic tempo, and the trial-specific scene configuration. LOSO on JIGSAWS can therefore be solved by surgeon re-identification followed by a table lookup, without any transferable notion of skill, whereas LOUO removes precisely this shortcut: the test surgeon appears in no training fold, so the model must generalise across operators. The generalisation claims of this paper consequently rest on LOUO alone.

The result is also not unprecedented. Table~\ref{tableSOTA1-2} shows that the 3D-convolutional baseline reports $100\%$ LOSO accuracy on Suturing and Needle-Passing and $95.8\%$ on Knot-Tying; saturation under this protocol is a documented property of the benchmark, which is a further reason to read a perfect LOSO score as a consistency check rather than as evidence of superiority.

Three further considerations explain why all feature strategies yield identical numbers. First, the metric has coarse resolution and a zero-error state is absorbing: with the small number of test videos available in each LOSO fold, accuracy moves in discrete steps within a fold, and once every configuration classifies every test video correctly the metric can no longer distinguish them; identical values therefore indicate that all configurations lie above the separation threshold, not that they are the same model. Second, the configurations share everything except the kernel composition: all twelve use the same RF features, the same tree and the same inference procedure, and differ only in which sub-kernels are summed, so when the video-level representations of the classes are already well separated in the training fold every composition yields the same zero-error decision rule. Third, the configurations are demonstrably \emph{not} equivalent when the problem is not saturated: on Capsulorhexis, where several configurations tie at $0.892$ accuracy, the AUC still separates them across the range $0.862$--$0.974$, and under LOUO the same twelve configurations spread over $22$ accuracy points ($75.0$--$97.3$). Consistently with this reading, the standard deviation of the LOSO accuracy across the five folds is exactly zero for every configuration, which is itself informative: a zero dispersion over five independent folds reflects the saturation described above rather than an unusually stable model.

Compared with the more challenging LOUO protocol, these results confirm that LOSO constitutes a less demanding evaluation setting, where the proposed RF-DUAGP-Tree framework can reliably distinguish between surgical skill levels regardless of the selected feature configuration or SCUK formulation.

\subsection{Ablation Analysis}
\label{A-A}
The ablation analysis was conducted to investigate the contributions of different feature-utilization strategies and uncertainty-aware kernel formulations within the proposed RF-DUAGP-Tree framework. The analysis is based on the experimental results reported in Tables~\ref{table3}--\ref{table9}.

The effect of each component, computed from the results tables, is summarised in Table~\ref{tabAblation}. The joint use of the two RF feature streams is by far the dominant factor (up to $+22.3$ accuracy points under LOUO); the inclusion of the higher-order dynamic (jerk) term is a consistent but much smaller second contributor ($0$ to $+3.5$ points, and $+0.056$ AUC on Capsulorhexis); and the choice among the three SCUK formulations is the smallest factor, improving on the reduced kernel of Eq.~\eqref{kav} but not on the full composition of Eq.~\eqref{kx}. This finding suggests that compact statistical representations are more suitable for limited-data surgical skill assessment scenarios.
\begin{table}
\centering
\caption{Contribution of each component, computed from Tables~\ref{table3},~\ref{table4},~\ref{table6} and~\ref{table8}}
\label{tabAblation}
\begin{tabular}{@{}p{0.30\textwidth} p{0.30\textwidth} p{0.34\textwidth}@{}}
\toprule
Component & Comparison & Effect (accuracy points) \\
\midrule
\textbf{1. Joint use of both RF feature streams} & Strategy~1 vs.\ Strategy~4 (single stream), JIGSAWS LOUO & \textbf{+21.2} (Sut.), \textbf{+22.3} (KT), \textbf{+22.2} (NP) \\
 & Strategy~1 vs.\ Strategy~3, JIGSAWS LOUO & +8.7, +9.8, +9.7 \\
 & Strategy~1 vs.\ Strategy~4, Capsulorhexis & $0.0$ (0.892 vs.\ 0.892) \\
\textbf{2. Dynamic features (jerk term)} & SCUK1 Eq.~\eqref{kx} vs.\ SCUK2 Eq.~\eqref{kav}, JIGSAWS LOUO & +2.8 (Sut.), $0.0$ (KT), +1.4 (NP) \\
 & SCUK1 vs.\ SCUK2, Capsulorhexis & +3.5 accuracy; +0.056 AUC (0.974 vs.\ 0.918) \\
\textbf{3. Entropy-/uncertainty-aware modulation} & SCUK3 Eq.~\eqref{kj} vs.\ SCUK2, JIGSAWS LOUO & +2.8 (Sut.), $0.0$ (KT), +0.8 (NP) \\
 & SCUK3 vs.\ SCUK1, JIGSAWS LOUO & $0.0$ (Sut.), $0.0$ (KT), $-0.6$ (NP) \\
 & SCUK3 vs.\ SCUK1, Capsulorhexis & $0.0$ accuracy; $-0.031$ AUC \\
\bottomrule
\end{tabular}
\end{table}

The Capsulorhexis results further highlight the importance of the proposed feature summarization strategy. Empirically, utilizing the original feature maps without abstraction resulted in lower classification performance, whereas the summarized feature representation consistently improved the effectiveness of the DUAGP-Tree classifier.

The results consistently demonstrate the importance of the feature-utilization strategy. Across all JIGSAWS tasks under the more challenging LOUO protocol, Strategy 1 achieves the highest classification performance. This strategy jointly utilizes the flow-layer and pre-classification-layer feature streams, indicating that the two representations provide complementary information for surgical skill assessment. In contrast, strategies relying on partial feature combinations or individual feature streams generally result in lower classification performance.

To quantify how much of the reported performance depends on the particular kernel choice, the accuracy dispersion across the three SCUK variants within Strategy 1 under LOUO is $95.3 \pm 1.6$ (Suturing), $97.3 \pm 0.0$ (Knot-Tying), and $96.5 \pm 0.7$ (Needle-Passing). Because each LOUO fold contains only $3$--$5$ test trials, i.e.\ $6$--$10$ test videos once both camera views are counted, a single misclassified video moves the reported accuracy by $\approx 1$--$3$ points, and differences of this order between configurations should not be over-interpreted.

The analysis further highlights the contribution of the proposed uncertainty-aware kernel design. The three uncertainty-aware formulations of Eqs.~\eqref{UA1rbf}--\eqref{EArbf} are complementary options that may be assigned to any of the semantic sub-kernels, and all three were used across the reported experiments. Among the configurations examined, those in which the acceleration or the jerk sub-kernel took the EA form were consistently among the best performing and, more usefully, the most stable: the entropy-of-entropy quantity supplies a well-scaled initial value for the ARD length-scales of the higher-order dynamic blocks, which lowers the sensitivity of the fit to the remaining hyperparameters and removes the need to re-tune initial values between tasks. This is reported as a practical default rather than as a ranking of the three formulations against one another. Overall, the results confirm that both the proposed feature integration strategy and the uncertainty-aware kernel formulation contribute to the effectiveness of the RF-DUAGP-Tree framework.

\subsection{Comparison with State-of-the-Art Methods}

The state-of-the-art comparison includes an RF-GP-Tree (DKL) baseline to provide a video-compatible reference for the proposed classifier. The original GP-Tree framework was introduced for image classification~\cite{achituve2021gp}; for the present video-based task, we use Representation Flow (RF) as the feature extractor and retain the DKL-based GP-Tree classification stage. This study-specific baseline is denoted as RF-GP-Tree (DKL), and it is evaluated on the same datasets and corresponding video-level partitions as RF-DUAGP-Tree: LOSO and LOUO for JIGSAWS, and the same Capsulorhexis protocol of three repeated training runs, each with five test rounds of 28 videos following training on 30 videos. The published 3D ConvNet, VBA-Net, TimeSformer and R3D18 entries are included as external references and are retained as reported in their cited sources. Thus, the RF-GP-Tree (DKL) rows should be distinguished from results copied from prior benchmark papers.

Under JIGSAWS LOSO, RF-DUAGP-Tree reaches 100\% on suturing, knot tying and needle passing, while the RF-GP-Tree (DKL) baseline is reported at 94--98\%, 94--98\% and 95--98\%, respectively (Table~\ref{tableSOTA1-2}). These results are presented for completeness and protocol-level comparison. LOSO is not a cross-subject protocol: every surgeon appears in the training set of each fold, and the skill label is constant for a given surgeon. Moreover, the high results of several methods indicate that this protocol can be saturated; therefore, the LOSO results should not be interpreted as evidence of generalisation to unseen surgeons.

The more informative JIGSAWS comparison for subject-independent evaluation is LOUO (Table~\ref{tableSOTA1}). RF-DUAGP-Tree obtains 96.2\% on suturing, 97.3\% on knot tying and 97.2\% on needle passing, yielding a mean accuracy of 96.90\%. The corresponding RF-GP-Tree (DKL) entries are 92--95\%, 91--93\% and 91--93\%, with a reported mean range of 91--93.66\%. The proposed method is above the upper end of the listed RF-GP-Tree (DKL) range for each task. The comparison with VBA-Net is also favourable for the proposed method across the three task-wise ranges shown. However, the 3D ConvNet ranges overlap with the proposed results on suturing and needle passing, so the table does not support a claim that RF-DUAGP-Tree outperforms every published method on every task. The main generalisation evidence is therefore the LOUO evaluation and the comparison with the video-compatible RF-GP-Tree (DKL) baseline, interpreted in light of the fold-wise ranges.

On Capsulorhexis (Table~\ref{tableSOTA2}), the reported mean performance of RF-DUAGP-Tree across all 15 test rounds (five rounds in each of three repeated training runs) is 89.2\% accuracy, 91.1\% precision, 89.2\% recall and 89.1\% F1-score. Compared with the corresponding mean metrics for RF-GP-Tree (DKL), these values represent gains of 3.9, 2.8, 3.9 and 3.8 percentage points, respectively. Among the reported RF-DUAGP-Tree configurations, M3 (Strategy 1, SCUK3) provides the highest precision and F1-score, while M1 (Strategy 1, SCUK1) achieves the same accuracy and the highest AUC (0.974); the full set of configurations is reported in Table~\ref{table3}. The TimeSformer and R3D18 results are the published Cataract-LMM benchmark results and were obtained using approximately 140 training videos, whereas each training run of the proposed approach uses 30 training videos and each testing round uses 28 test videos. The reported 89.2\% accuracy summarizes the 15 test rounds and is not the score of a single uniquely fixed 28-video test subset.

Taken together, Tables~\ref{tableSOTA1-2}--\ref{tableSOTA2} place the proposed method alongside both published video-based systems and a DKL-based GP-Tree adapted to the same video modality. Because the RF-GP-Tree (DKL) baseline and RF-DUAGP-Tree share the RF video feature-extraction backbone and use matching dataset-specific evaluation protocols—including the same Capsulorhexis protocol of three runs with five 28-video test rounds per run—the comparison provides a more direct reference for the classifier designs than a comparison between unrelated backbones. Nevertheless, the proposed method combines dynamic descriptors and uncertainty-aware kernel changes; the tables therefore evaluate the overall proposed design and do not isolate the individual contribution of each modification.

\begin{table}[!th]
\centering
\caption{Comparison with state-of-the-art methods on JIGSAWS under the LOSO protocol.}
\label{tableSOTA1-2}
\begin{tabular*}{\textwidth}{@{\extracolsep\fill}lcccccl}

\toprule%
Method & Year & Modality& Suturing & Knot Tying & Needle Passing & Mean\\

\midrule
%Methods & Acc & F1.s & Rec & Acc & F1.s & Rec & Acc & F1.s & Rec\\
%\midrule

3D ConvNet (Funke et al.)~\cite{funke2019video}  & 2019 & video (RGB+OF)& 100 & 95.8 & 100 & 95-100\\
VBA-Net (Yanik et al.)~\cite{yanik2023video} & 2023 & video & 93-95 & 92-94 & 92-95 & 93-95\\
RF-GP-Tree (DKL)$^{*}$ & 2021 & video & 94-98 & 94-98 & 95-98 & 94-98\\
RF-DUAGP-Tree (Ours) & 2026 & video & 100 & 100 & 100 & 100\\
\bottomrule
\end{tabular*}
\caption*{The RF-DUAGP-Tree row reports the best-performing configuration among the twelve evaluated configurations (M1: Strategy 1, SCUK1); the complete grid is reported in Tables~\ref{table5},~\ref{table7}, and~\ref{table9}. The RF-GP-Tree (DKL) row is a video-compatible baseline evaluated in this study: it uses Representation Flow as the feature extractor and retains the DKL-based GP-Tree classifier, with the same LOSO video-level partitions used for the proposed method. The year (2021) refers to the original GP-Tree publication~\cite{achituve2021gp}; the asterisk marks the video-compatible adaptation evaluated in this study, not a rerun of the original image-classification pipeline unchanged. The 3D ConvNet and VBA-Net entries are taken from the cited publications; their reported fold ranges and aggregate values are reproduced as reported rather than recomputed from the endpoints displayed in the task-wise columns. LOSO is included for comparability only: because each surgeon occurs in the training folds and the skill label is constant per surgeon, it does not establish cross-subject generalisation. The main subject-independent results are in Table~\ref{tableSOTA1}.}
\end{table}

\begin{table}[!th]
\centering
\caption{Comparison with state-of-the-art methods on JIGSAWS under the LOUO protocol.}
\label{tableSOTA1}
\begin{tabular*}{\textwidth}{@{\extracolsep\fill}lcccccl}

\toprule%
Method & Year & Modality& Suturing & Knot Tying & Needle Passing & Mean\\

\midrule
%Methods & Acc & F1.s & Rec & Acc & F1.s & Rec & Acc & F1.s & Rec\\
%\midrule

3D ConvNet (Funke et al.)~\cite{funke2019video}   & 2019 & video (RGB+OF)& 94-100& 86-95 & 90-100 & 95-100\\
VBA-Net (Yanik et al.)~\cite{yanik2023video} & 2023 & video & 93-95 & 92-94 & 93-95 & 94-95\\
RF-GP-Tree (DKL)$^{*}$ & 2021 & video & 92-95 & 91-93 & 91-93 & 91-93.66\\
RF-DUAGP-Tree (Ours) & 2026 & video & 96.2 & 97.3 & 97.2 & 96.90\\
\bottomrule
\end{tabular*}
\caption*{The RF-DUAGP-Tree row reports the best-performing configuration among the twelve evaluated configurations (M1: Strategy 1, SCUK1); the complete grid is reported in Tables~\ref{table4},~\ref{table6}, and~\ref{table8}. The RF-GP-Tree (DKL) row is the video-compatible baseline evaluated in this study using the RF feature extractor and DKL-based GP-Tree classifier under the same LOUO video-level partitions. The year (2021) refers to the original GP-Tree publication~\cite{achituve2021gp}; the asterisk marks the video adaptation evaluated in this study. The 3D ConvNet and VBA-Net values are taken from their cited publications; reported ranges/aggregate values are retained as published rather than recalculated from the endpoints of the task-wise ranges. The comparison with 3D ConvNet is mixed because its reported ranges overlap with RF-DUAGP-Tree for suturing and needle passing; therefore, no across-the-board superiority over every published method is claimed.}
\end{table}

\begin{table}[!th]
\centering
\caption{Comparison with state-of-the-art methods on the Capsulorhexis dataset.}
\label{tableSOTA2}
\begin{tabular}{lccccc}
\toprule
Method & Accuracy & Precision & Recall & F1-Score\\
\midrule
TimeSformer~\cite{2026cataract} & 82.5 & 86 & 82 & 83.90\\
R3D18~\cite{2026cataract} & 81.67 & 82.35 & 84.85 & 83.58\\
RF-GP-Tree (DKL)$^{*}$ & 85.3 & 88.3 & 85.3 & 85.3\\
\textbf{RF-DUAGP-Tree (Ours)} & \textbf{89.2} & \textbf{91.1}& \textbf{89.2}& \textbf{89.1}\\
\bottomrule
\end{tabular}
\caption*{The RF-DUAGP-Tree row reports M3 (Strategy 1, SCUK3) from the twelve configurations in Table~\ref{table3}, which attains the highest precision and F1-score; M1 (Strategy 1, SCUK1) reaches the same accuracy and the highest AUC (0.974). The complete grid is reported in Table~\ref{table3}. RF-GP-Tree (DKL)$^{*}$ is the video-compatible baseline evaluated in this study using the RF feature extractor and DKL-based classifier under the same Capsulorhexis protocol of three repeated runs, each comprising five 28-video test rounds following training on 30 videos. The RF-GP-Tree (DKL) and RF-DUAGP-Tree entries report mean metrics across all 15 test rounds (five in each of three repeated runs); the published TimeSformer and R3D18 entries are external benchmark results. The asterisk indicates that the original GP-Tree method~\cite{achituve2021gp} was adapted for video input, and that the results shown are from this study rather than from the original image-classification experiments. In contrast, the TimeSformer and R3D18 values are taken from the published Cataract-LMM benchmark~\cite{2026cataract}, whose deep baselines used approximately 140 training videos. The values in this table are percentages.}
\end{table}
\subsection{Model capacity, complexity and runtime}
\label{modelcapcity}
This section reports the quantities that support the efficiency statements made in this paper: the number of parameters that are actually adapted, the asymptotic complexity of the classifier, and the training, inference and memory cost measured on a single, stated hardware configuration. They are collected in Table~\ref{tablemodelcapcity}.

Three or four of the five convolutional blocks of the backbone are fine-tuned for three epochs, together with the flow-layer scalars $\lambda$, $\theta$ and $\tau$, and the classifier that is fitted on top of them has $20{,}529$ trainable parameters on JIGSAWS and $31$ on Capsulorhexis ($20{,}530$ and $35$ when the entropy-modulated composition of Eq.~\eqref{kj} is used). Fitting DUAGP-Tree costs $O\big((C{-}1)(Nm^2+m^3)\big)$ per training iteration and $O\big((C{-}1)m^2\big)$ per test video, against $O(N^3)$ for a full multi-class Gaussian process classifier; in wall-clock terms this is approximately $45$\,s per fold on JIGSAWS and $10$\,s on Capsulorhexis, against $40$--$90$\,min per fold for the Representation Flow adaptation. Two cost properties are structural rather than measured: no optical flow is pre-computed at any stage, which in a conventional two-stream pipeline requires a full TV-L1 pass over every frame pair of every video before training begins~\cite{funke2019video, piergiovanni2019representation}; and at inference the backbone is executed once per video in inference mode, with no back-propagation, so the per-video classification cost after feature extraction is on the order of a millisecond. All measurements were obtained on a single laptop-class GPU with $4$\,GB of memory, which is itself a statement about the resource envelope the method requires.
\begin{table}[!th]
\centering
\caption{A report of parameters of the structures}
\label{tablemodelcapcity}
\begin{tabular}{@{}p{0.36\textwidth} p{0.28\textwidth} p{0.28\textwidth}@{}}
\toprule
Quantity & JIGSAWS & Capsulorhexis \\
\midrule
Backbone blocks adapted (of five) & 4 or 3, fine-tuned for 3 epochs & 3, fine-tuned for 3 epochs \\
Classifier trainable parameters & $20{,}529$ ($20{,}530$ with $k_{entropy}$) & $31$ ($35$ with $k_{entropy}$) \\
Training videos per fold & $\approx68$ & $30$ \\
Feature extraction, per video & $\approx11$\,s & $\approx11$\,s \\
RF adaptation, per fold & $40$--$90$\,min & $40$--$90$\,min \\
DUAGP-Tree training, per fold & $\approx45$\,s & $\approx10$\,s \\
Inference, per video (excl.\ feature extraction) & $\approx1$\,ms & $\approx1$\,ms \\
Peak GPU memory & $\le4$\,GB & $\le4$\,GB \\
Classifier complexity & \multicolumn{2}{@{}p{0.58\textwidth}@{}}{$O\!\big((C{-}1)(Nm^2 + m^3)\big)$ per training iteration; $O\!\big((C{-}1)m^2\big)$ per test video} \\
Hardware & \multicolumn{2}{@{}p{0.58\textwidth}@{}}{Intel i7-6700HQ, 12\,GB RAM, NVIDIA GTX 960M (4\,GB VRAM)} \\
Software & \multicolumn{2}{@{}p{0.58\textwidth}@{}}{Ubuntu 24.04.4, Python 3.10.12, CUDA 12.2, PyTorch 2.7.1+cu118, GPyTorch 1.15.2} \\
\bottomrule
\end{tabular}
\caption*{Classifier parameter counts are obtained by enumerating the parameters of the implementation. On Capsulorhexis the single GP node carries $27$ ARD length-scales ($5+8+8+6$, one per input feature) together with the four composition weights $\alpha,\beta,\gamma,\delta$. On JIGSAWS each of the two GP nodes carries $10{,}259$ parameters, dominated by the $4\times512+4\times2048=10{,}240$ ARD length-scales of the eight sub-kernels of the two feature streams; together with the stream and semantic composition weights $a,b,\alpha,\beta,\gamma,\delta$ this gives $20{,}529$ in total. The entropy kernel of Eq.~\eqref{kj} contributes one additional parameter on JIGSAWS and four on Capsulorhexis. For comparison, the R3D18 and TimeSformer baselines of Table~\ref{tableSOTA2} are trained end-to-end and their parameter counts are three orders of magnitude larger.}
\end{table}

\begin{figure*}
\centering
\subcaptionbox[]{\label{figLouJKN1}}{\includegraphics[width=6cm]{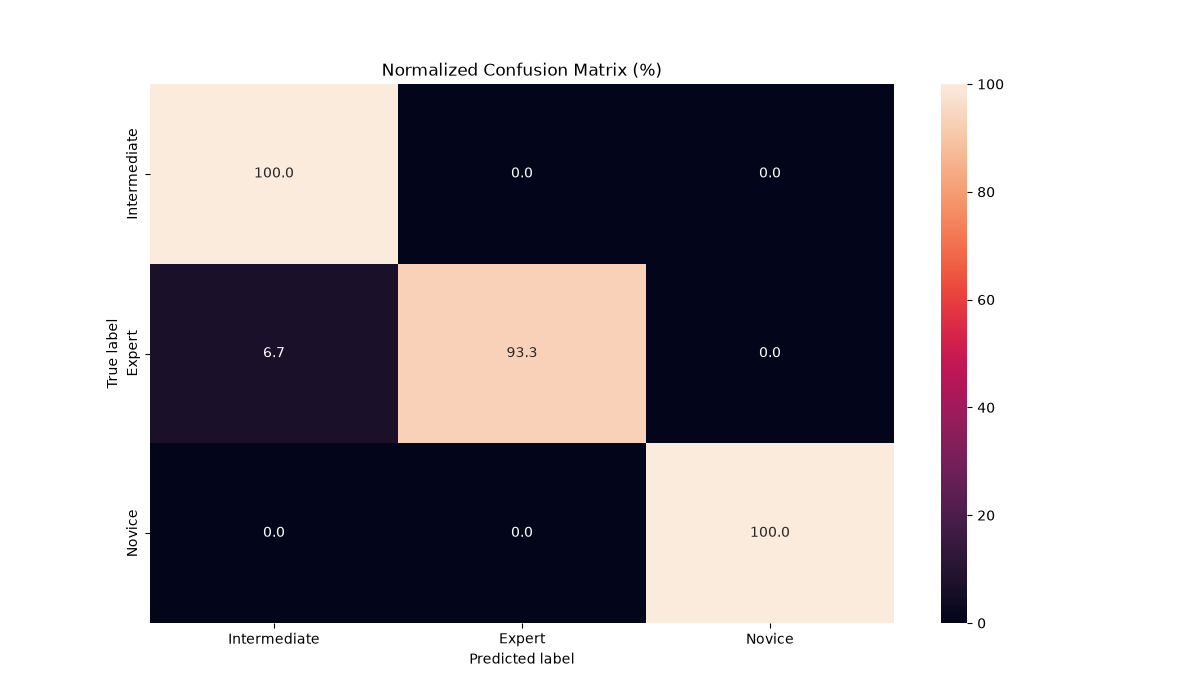}}%
\hfill
\subcaptionbox[]{\label{figLouNE2}}{\includegraphics[width=6cm]{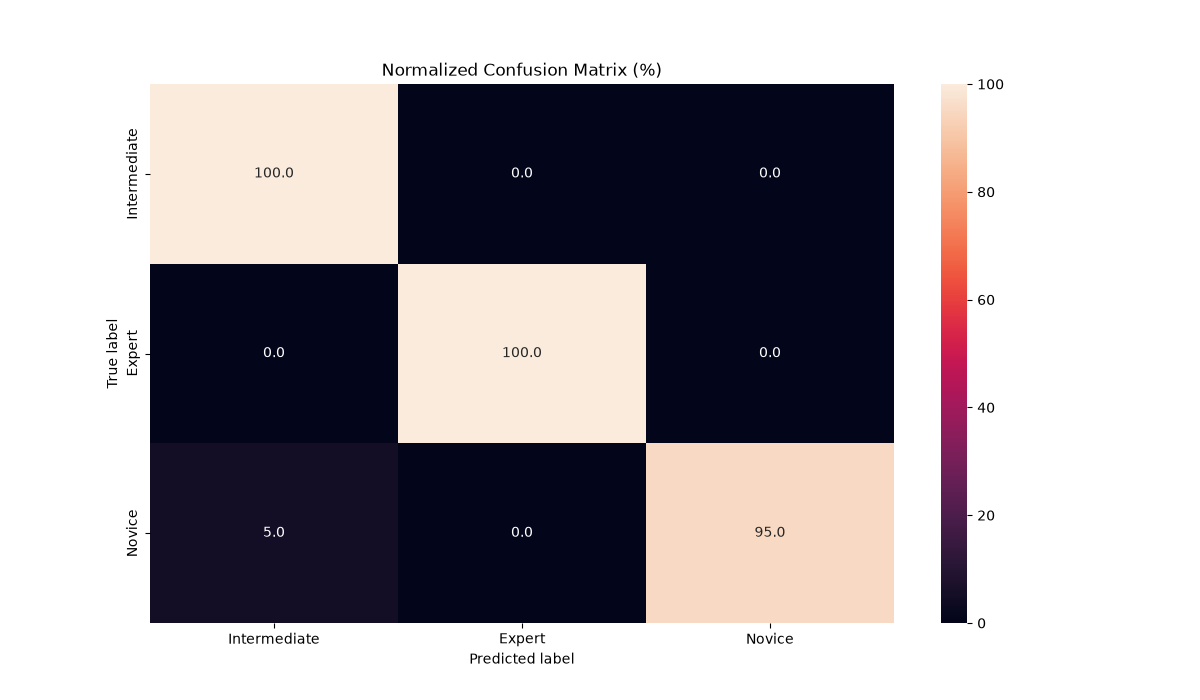}}%
\subcaptionbox[]{\label{figJLouST2}}{\includegraphics[width=6cm]{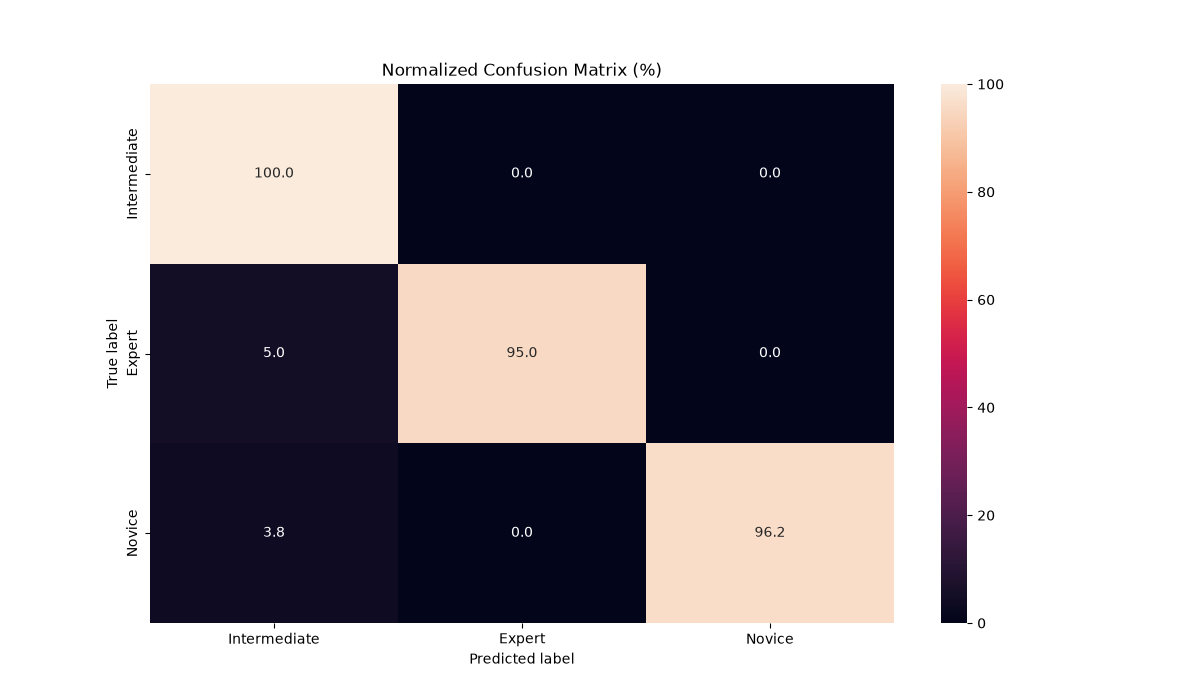}}%

\subcaptionbox[]{\label{figLosJKN1}}{\includegraphics[width=6cm]{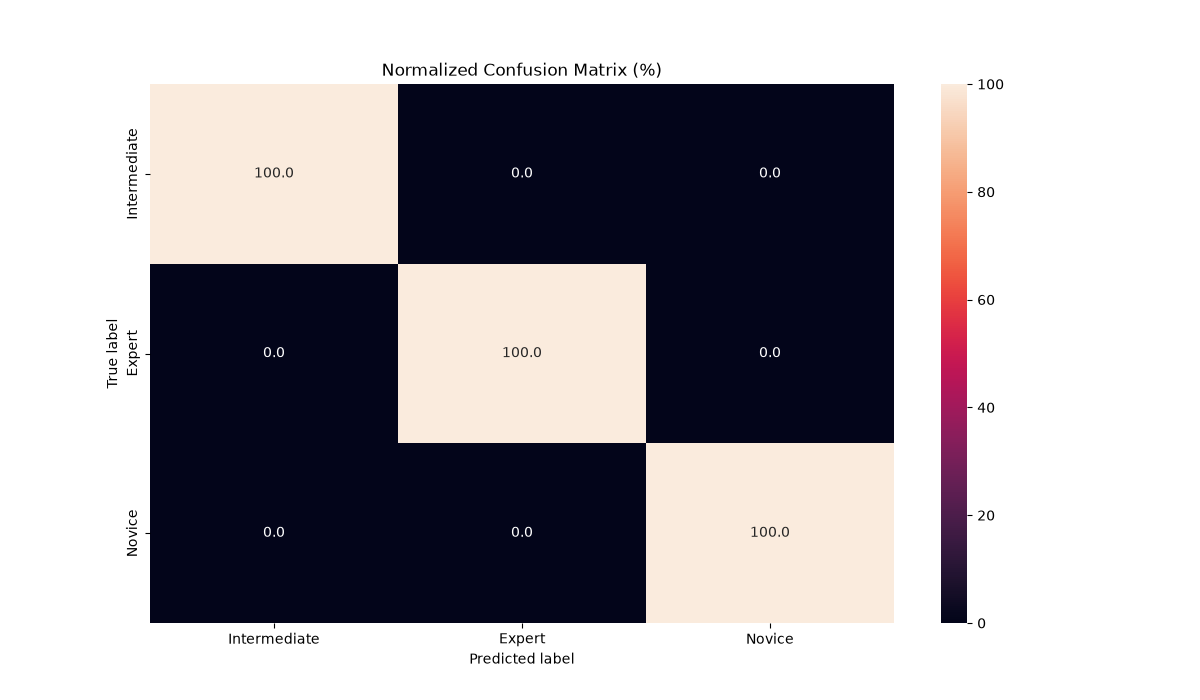}}%
\hfill
\subcaptionbox[]{\label{figLosNE2}}{\includegraphics[width=6cm]{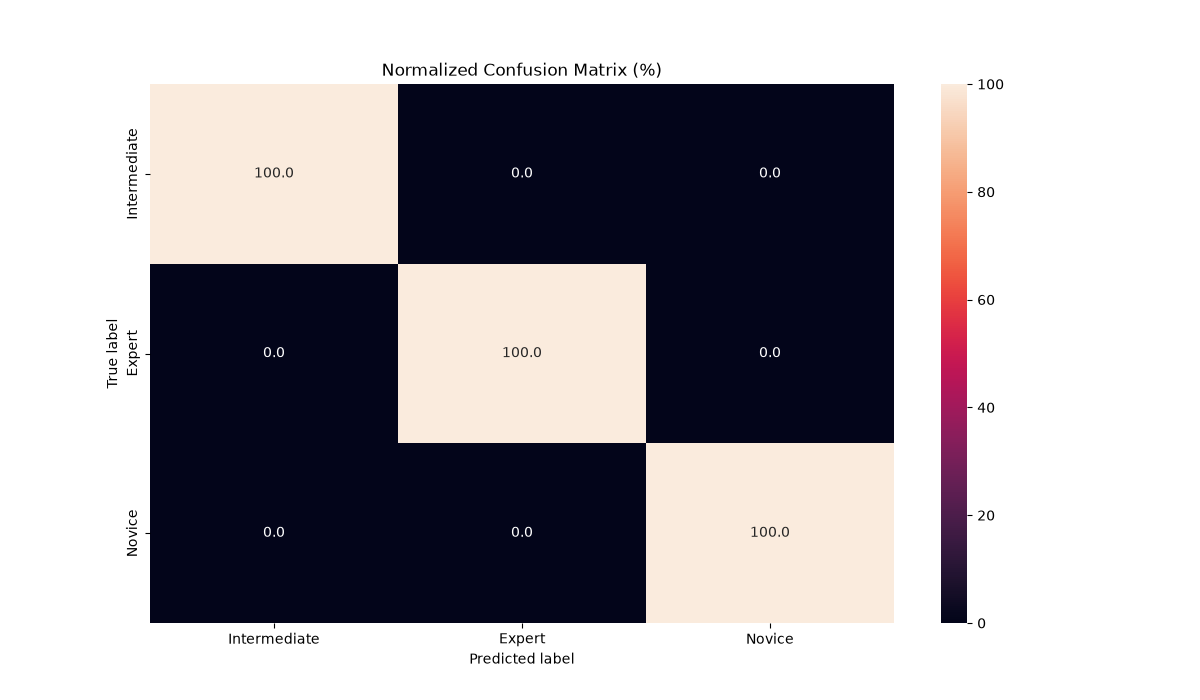}}%
\subcaptionbox[]{\label{figJLosST2}}{\includegraphics[width=6cm]{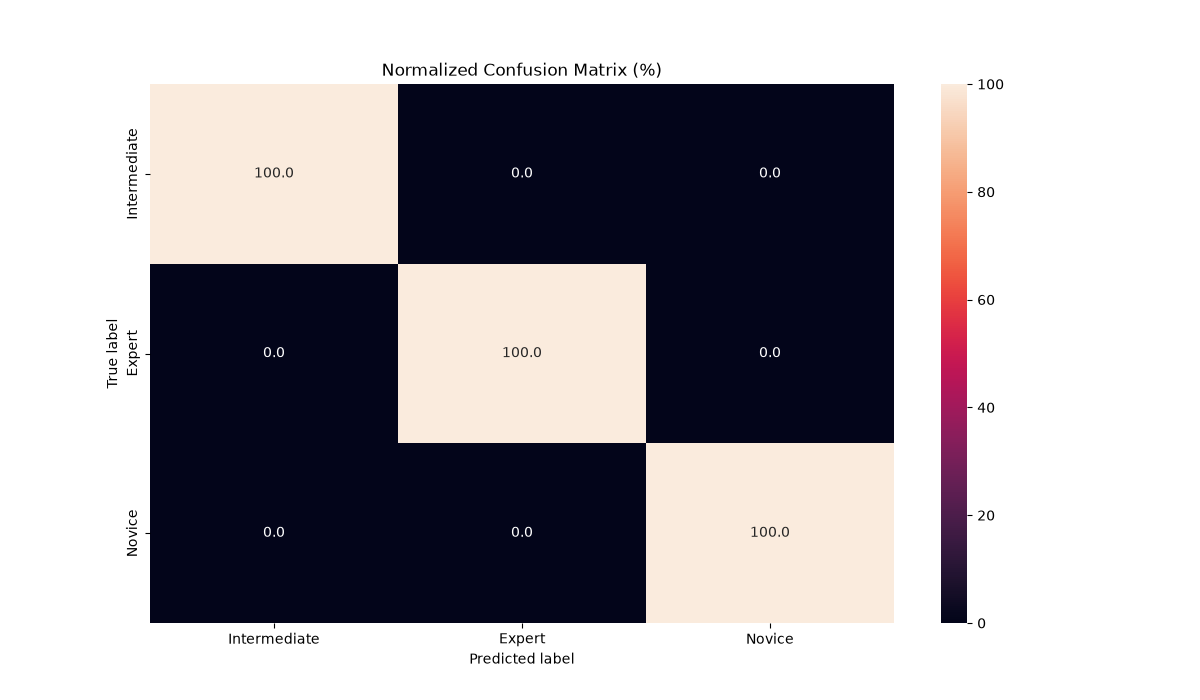}}%

\caption{Confusion matrices for the reported configuration. (a) Knot-Tying under the LOUO protocol; (b) Needle-Passing under the LOUO protocol; (c) Suturing under the LOUO protocol; (d) Knot-Tying under the LOSO protocol; (e) Needle-Passing under the LOSO protocol; (f) Suturing under the LOSO protocol.}
 \label{JIGS}
\end{figure*}

\begin{figure*} [!ht]
\centering
\includegraphics[width=6cm]{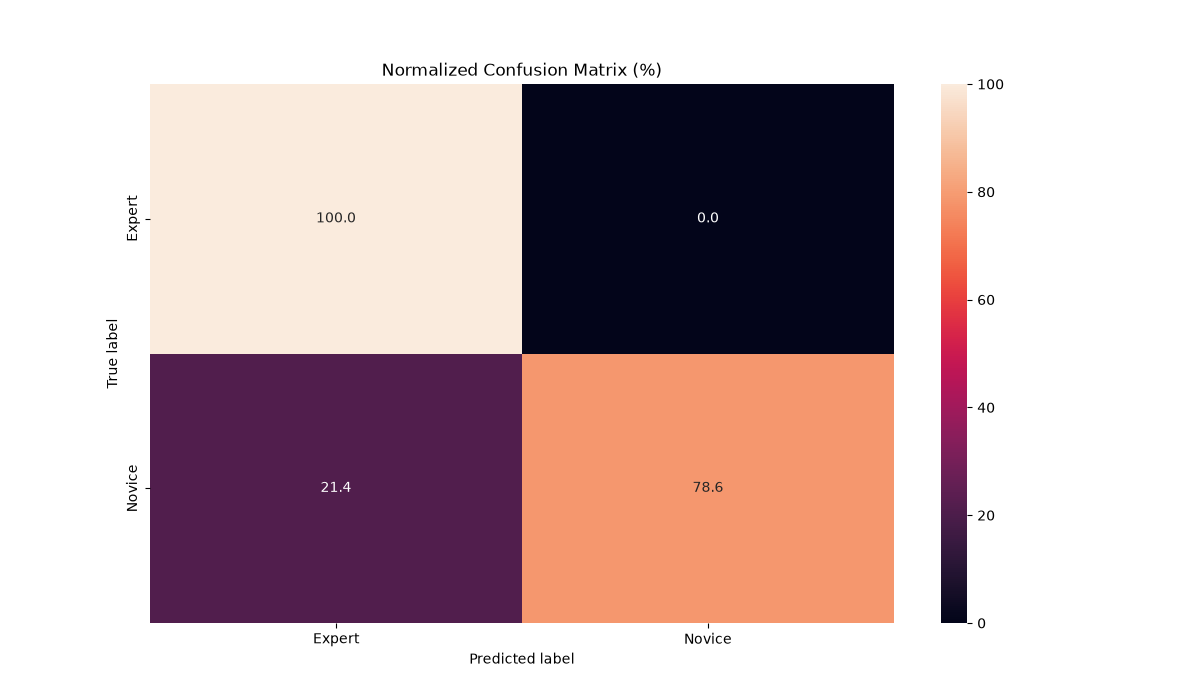}
\caption{Confusion matrix for the Capsulorhexis dataset under the repeated evaluation protocol: three training runs, each with five test rounds of 28 videos.}
  \label{CAPS}
\end{figure*}

The corresponding confusion matrices for the JIGSAWS and Capsulorhexis datasets are presented in Fig.~\ref{JIGS} and Fig.~\ref{CAPS}, providing a detailed visualization of the classification performance across the respective skill levels.

\section{Discussion}
\label{Discussion}

\subsection{Summary of contributions and main findings}
The main contributions of this study are fourfold. First, we propose RF-DUAGP-Tree, a novel framework that integrates Representation Flow-based spatio-temporal feature extraction with a dynamics- and uncertainty-aware Gaussian Process Tree classifier for surgical skill assessment. Second, we introduce a set of clip-level semantic motion descriptors, including motion trend, acceleration, and jerk, which capture higher-order latent surgical dynamics and serve both as discriminative features and uncertainty indicators. Third, we develop Semantic Compositional Uncertainty Kernels (SCUK), a structured kernel design that explicitly models the heterogeneous semantics of latent motion representations. In the two datasets studied here, the dataset-dependent choices are representation compactness and stream composition. Finally, we propose uncertainty-aware ARD-RBF kernels that incorporate entropy-driven measures of latent motion variability, improving robustness to temporal fluctuations, abnormal transitions, and limited-data training scenarios while keeping the number of adapted parameters small (Section~\ref{modelcapcity}).

\subsection{Implications for surgical training and quality assurance}
A practical implication of this work is the feasibility of deploying video-only assessment in settings where kinematic or tool-tracking signals are unavailable or costly to obtain. A lightweight model that operates on limited  data can reduce the burden of full-length video processing and can support more frequent formative assessment. In training curricula, such a system can be integrated as (i) a periodic proficiency check, (ii) an objective progress indicator across sessions, and (iii) a triage tool to direct expert feedback toward trainees most in need of supervision. From a quality-assurance perspective, the low inference cost of the DUAGP-Tree-based classifier (Section~\ref{modelcapcity}) supports near-real-time use, which may enable scalable auditing in simulation centers. These are prospective applications. No user study, questionnaire or structured evaluation with surgeons was conducted as part of this work, and the clinical utility of the framework --- including whether its outputs are interpretable and actionable for trainees --- remains to be established. We regard a prospective evaluation with surgical educators as a necessary next step before any deployment claim can be made.

\subsection{Positioning with respect to prior video-based SSA literature}
Prior work on JIGSAWS has demonstrated that spatiotemporal video models (e.g., 3D CNNs) can achieve strong performance but often require substantial compute and/or a lot of videos for processing. In contrast, our approach targets the regime of limited data and emphasizes computational efficiency (Table~\ref{tablemodelcapcity}) by combining a pretrained motion-aware backbone (Representation Flow) with a sparse non-parametric Bayesian strategy. Importantly, the GP-based classifier provides a probabilistic decision function and can be trained with comparatively few trainable parameters, which is beneficial on datasets with limited trials per class such as Capsulorhexis (Table~\ref{table3}).

\subsection{Limitations and future work}
\label{limits}
The following limitations should be taken into account when interpreting the results.\\[4pt]
\emph{(i) Evaluation scope.} The framework is evaluated on two datasets: JIGSAWS, comprising three bench-top tasks with three skill levels, and the Capsulorhexis subset of Cataract-LMM, comprising one clinical task with two skill levels. JIGSAWS is a simulator dataset recorded at a single centre with a fixed camera; generalisation to other procedures, camera viewpoints, and clinical environments therefore remains to be validated. No cross-dataset transfer experiment was performed: a model trained on one dataset was not evaluated on the other.\\[4pt]
\emph{(ii) Model selection.} We report the complete grid of twelve feature and kernel configurations for every task and protocol, so that no configuration is hidden; however, no nested cross-validation or inner validation loop was used to select among them. The RF-DUAGP-Tree entries in Tables~\ref{tableSOTA1-2}--\ref{tableSOTA2} report the selected best-performing configurations from the fully reported grid and should therefore be read as an upper estimate of achievable performance rather than as an unbiased estimate of the generalisation error of a pre-specified configuration. A less biased estimate would require nested cross-validation, which we identify as necessary future work. The published 3D ConvNet, VBA-Net, TimeSformer and R3D18 entries are external reference results, while RF-GP-Tree (DKL) is the video-compatible baseline evaluated in this study; these rows are not part of the RF-DUAGP-Tree configuration grid.\\[4pt]
\emph{(iii) Stochastic variability and partition averaging.} The model runs use the single recorded random seed released with the code; experiments with multiple independent model-initialisation seeds were not performed. The stochastic surface is not empty: the backbone is fine-tuned for three epochs, the kernel hyperparameters are updated by a gradient-based optimiser, and the tree construction and inducing-point initialisation use $k$-means++. JIGSAWS results are reported over the standard cross-validation folds and configurations. For Capsulorhexis, the reported metrics are averages over 15 test rounds, with five rounds in each of three repeated training runs; each run used 30 training videos and each test round used 28 test videos.\\[4pt]
\emph{(iv) Computational reporting.} We report trainable parameter counts, asymptotic complexity, and measured training and inference times on a single stated hardware configuration (Table~\ref{tablemodelcapcity}). We did not profile floating-point operations or rerun R3D-18 and TimeSformer on our hardware; their entries in Table~\ref{tableSOTA2} are taken from the published Cataract-LMM benchmark. We did, however, evaluate a video-compatible RF-GP-Tree (DKL) baseline for Tables~\ref{tableSOTA1-2}--\ref{tableSOTA2}, pairing the RF feature extractor with the DKL-based GP-Tree classifier and using the corresponding data partitions described in Section~\ref{Results}. This baseline is distinct from rerunning the original image-classification GP-Tree pipeline with its original image feature extractor. We therefore make no wall-clock superiority claim over R3D-18, TimeSformer, or the original image-based GP-Tree implementation. The RF-GP-Tree (DKL) results provide a modality-matched reference for the proposed video pipeline, but the comparison should be understood as a comparison of the complete configurations, not as an isolated test of DKL removal or uncertainty modeling alone. \\[4pt]
\emph{(v) Dataset-specific instantiation.} The pipeline is applied through two documented instantiations, one structured and one compact (Sections~\ref{DUAGPT} and~\ref{FrameworktoNewd}). The selection heuristic is based on these two datasets and has not been validated on a third. In particular, the compact representation for Capsulorhexis was chosen after the structured representation had underperformed, so this choice was informed by observed results rather than established prospectively through an inner validation procedure. Evaluation on an independent third dataset, ideally from a different surgical procedure and acquisition setting, is an important next step. The automatic tree construction has been exercised only on two- and three-class problems.\\[4pt]
\emph{(vi) Clinical evaluation.} No user study, questionnaire or structured evaluation with surgeons was conducted. Whether the framework's outputs are interpretable and actionable in surgical education remains an open question and a necessary prerequisite for any deployment claim.\\[4pt]
Future work will address these points in order of importance: nested cross-validation and multi-seed repetition; a learning-curve analysis and a matched-hardware comparison against transformer-based video backbones; cross-dataset and cross-centre transfer; extension to continuous skill scoring rather than discrete labels; and a prospective evaluation with surgical educators.

\subsection{Applying the framework to a new dataset}
\label{FrameworktoNewd}
The framework contains a fixed core and a small dataset-dependent part, and it is useful to state which is which. Fixed by the method, and unchanged between the two datasets studied here, are: the Representation Flow backbone and its two taps; the semantic decomposition of the representation into $f$, $v$, $a$ and $j$, which is defined by derivative order rather than by the data; the finite-difference dynamics operators; the tree construction; the uncertainty formulation with $\lambda_v=\lambda_a=\lambda_j=1$; and the ARD-RBF sub-kernel family. Dataset-dependent are two binary decisions: whether the representation is retained in structured form or reduced to summary statistics, and whether the two feature streams are modelled by separate kernels composed through Eq.~\eqref{kfe}.\\[5pt]
For a new dataset, the following procedure can serve as an initial guideline. (1)~Define the sample unit as a complete video or case, and choose the grouping key --- case, surgeon or centre --- according to the generalisation being claimed; partition on that key before extracting clips. (2)~Measure $N_{\mathrm{train}}$, the number of classes, the number of videos per class and the range of the clip count; no model is trained at this stage. (3)~Extract Representation Flow features once with the pretrained backbone and record the raw dimensionality $D_{\mathrm{raw}}$. (4)~Consider the compact variant when $D_{\mathrm{raw}}/N_{\mathrm{train}}$ is large or acquisition is heterogeneous across centres or devices, and the structured variant when the training set is comparatively large and acquisition is standardised. (5)~Select the stream composition by comparing $K_{\mathrm{flow}}$, $K_{\mathrm{end}}$ and $aK_{\mathrm{flow}}+bK_{\mathrm{end}}$ on an inner split of the training fold, adopting the composition only when the gain exceeds the inner-fold dispersion; this is the only step that requires model fitting, and each fit is a sparse Gaussian process over a few tens of samples. (6)~Leave the remaining components fixed at their defaults: the semantic composition of Eq.~\eqref{kx}, the uncertainty weights $\lambda_v=\lambda_a=\lambda_j=1$, and the ARD-RBF sub-kernel family, with the number of inducing points scaled to the size of the training fold.\\[5pt]
The framework has not been evaluated on problems with more than three skill classes, on unedited full-length operative video, or in a cross-centre transfer setting, and the automatic tree construction has been exercised only on two- and three-class problems.

\section{Conclusion}
\label{Conclusion}
In this work, we introduced the RF-DUAGP-Tree framework for surgical skill assessment, integrating representation flow-based spatio-temporal feature learning with a dynamics- and uncertainty-aware Gaussian Process Tree classifier. By modeling higher-order motion derivatives and introducing semantic compositional kernels alongside entropy-driven uncertainty-aware ARD-RBF formulations, the proposed method captures fine-grained surgical motion characteristics in a structured probabilistic manner. Extensive evaluations on JIGSAWS and Capsulorhexis demonstrate competitive performance under limited data regimes, while keeping the number of adapted parameters small and requiring no pre-computed optical flow (Table~\ref{tablemodelcapcity}), highlighting the effectiveness of combining structured semantic modeling with uncertainty-aware kernel design for surgical video analysis.\\

\section*{Data availability}

The datasets analysed in this study are publicly available. The JIGSAWS (JHU-ISI Gesture and Skill Assessment Working Set) can be accessed from the primary Johns Hopkins Computational Interaction and Robotics Laboratory repository at \url{https://cirl.lcsr.jhu.edu/research/hmm/datasets/jigsaws_release/} or via Figshare at \url{https://figshare.com/s/1fed2f611b9eef3ecc6b}. The Capsulorhexis dataset (Skill Assessment subset) is derived from the Cataract-LMM benchmark, which is publicly released on Hugging Face under the repository \texttt{mjahmadi/Cataract-LMM} at \url{https://huggingface.co/datasets/mjahmadi/Cataract-LMM} (DOI: \url{https://doi.org/10.57967/hf/8673}). Both datasets contain fully de-identified recordings collected under institutional oversight. No new human-participant data were collected for this study, and all ethical considerations regarding the use of publicly available de-identified data were observed. The implementation of RF-DUAGP-Tree, together with the exact fold-definition files used for all experiments and the random seed, is available at the repository \url{https://github.com/areferezaee/RF-DUAGP-Tree}.

% \section*{Data availability}

% The dataset analysed in this study is the publicly available JIGSAWS (JHU-ISI Gesture and Skill Assessment Working Set). It can be accessed from the primary Johns Hopkins Computational Interaction and Robotics Laboratory repository at \url{https://cirl.lcsr.jhu.edu/research/hmm/datasets/jigsaws_release/}. As an alternative, the dataset has also been made available on Figshare at \url{https://figshare.com/s/1fed2f611b9eef3ecc6b}. JIGSAWS contains fully de-identified recordings collected under institutional oversight (see~\cite{gao2014jhu}). No new human-participant data were collected for this study, and all ethical considerations regarding the use of publicly available de-identified data were observed.

\section*{Declarations}

\noindent\textbf{Authors' contributions.}
A.R. and M.J.A. contributed to Conceptualization, Methodology, Software, Validation, Formal analysis, Visualization, and Writing – original draft. A.M. contributed to Investigation, Resources, Validation, and Writing – review \& editing. H.D.T. contributed to Supervision, Project administration, and Writing – review \& editing.
% \emph{Arefeh Rezaei \& Mohammad Javad Ahmadi}: Conceptualization, Methodology, Software, Validation, Formal analysis, Visualization, Writing – original draft. 
% \emph{Amir Molaei}: Investigation, Resources, Validation, Writing – review \& editing. 
% \emph{Hamid D. Taghirad}: Supervision, Project administration, Writing – review \& editing.

\noindent\textbf{Ethics statement.}
This study analyzed only the publicly available, fully de-identified JIGSAWS and Cataract-LMM datasets and involved no new data collection with human participants. According to institutional policy, analyses of publicly available de-identified data do not require ethics committee review. The original data for both benchmarks were collected under institutional oversight; see~\cite{gao2014jhu} and~\cite{2026cataract}.

% \noindent\textbf{Ethics statement.}
% This study analyzed only the publicly available, fully de-identified JIGSAWS dataset and involved no new data collection with human participants. According to institutional policy, analyses of publicly available de-identified data do not require ethics committee review. The original JIGSAWS data were collected under institutional oversight; see~\cite{gao2014jhu}.

\noindent\textbf{Acknowledgements.}
This work was supported in part by the Iran National Science Foundation (INSF) under Grant 4041171.

\noindent\textbf{Declaration of competing interests.}
The authors declare that they have no known competing financial interests or personal relationships that could have appeared to influence the work reported in this paper. The study reports a methodological pipeline evaluated on the publicly available JIGSAWS and Cataract-LMM datasets (see Sections 2.1 and 2.2) and lists only academic affiliations on the title page.

% \noindent\textbf{Declaration of competing interests.}
% The authors declare that they have no known competing financial interests or personal relationships that could have appeared to influence the work reported in this paper. The study reports a methodological pipeline evaluated on the publicly available JIGSAWS dataset (see Section 2.1 and Table 1) and lists only academic affiliations on the title page.

%\section*{Acknowledgments}
%This was was supported in part by......

%Bibliography
\bibliographystyle{unsrt}  
\bibliography{references}

\end{document}